\documentclass[a4paper,fleqn]{cas-dc}

\usepackage[numbers]{natbib}
\usepackage{setspace}
\usepackage{subcaption}
\usepackage{textcomp}
\hypersetup{hidelinks}

\begin{document}
\let\WriteBookmarks\relax

\shorttitle{Spectral Haar Binarizer}
\shortauthors{M. Romaszewski and P. G{\l}omb}

\title[mode=title]{Spectral Haar Binarizer: Compact Boolean Contrast Representations for Hyperspectral Classification}

\author[1]{Micha{\l} Romaszewski}[orcid=0000-0002-8227-929X]
\ead{mromaszewski@iitis.pl}
\cormark[1]

\author[1]{Przemys{\l}aw G{\l}omb}[orcid=0000-0002-0215-4674]
\ead{pglomb@iits.pl}

\affiliation[1]{organization={Institute of Theoretical and Applied Informatics, Polish Academy of Sciences},addressline={Bałtycka 5},postcode={44-100},city={Gliwice},country={Poland}}

\begin{abstract}
Tsetlin Machines offer an interpretable, hardware-friendly alternative to conventional machine learning models, but applying them to continuous hyperspectral measurements requires converting spectral signals into Boolean representations. Standard binarisation methods typically discretise each spectral band independently, potentially discarding important relationships between neighbouring wavelengths. In this paper, we introduce the Spectral Haar Binarizer (SHB), a contrast-based spectral representation method that constructs Boolean predicates from differences between selected spectral intervals. Inspired by Haar-like contrast features, the method uses an AdaBoost-style selection procedure to obtain a compact set of discriminative spectral predicates. We evaluate SHB on five heterogeneous hyperspectral datasets covering forensic blood detection, agricultural and urban remote sensing, and rainforest tree-species classification. Across binary and multiclass tasks, SHB consistently outperforms standard per-band binarisation strategies and provides the strongest or near-strongest results across classifiers. In binary tasks, TM-SHB achieves the best average performance, while RF-SHB is strongest in multiclass settings. Bayesian and rank-based analyses confirm that SHB's advantage is robust across datasets and not driven by isolated cases. The results show that representing continuous hyperspectral measurements is a central processing step for efficient classification. That contrast-based Boolean features provide an effective bridge between information-rich sensor measurements and compact, logic-based processing.
\end{abstract}

\begin{keywords}
Hyperspectral imaging \sep binarisation \sep Tsetlin machines \sep machine learning
\end{keywords}

\maketitle

\section{Introduction}
Although much of contemporary machine learning is driven by the pursuit of increasingly general models, reflecting the homogenising tendency of foundation models to serve as shared bases for diverse downstream tasks \cite{Bommasani2022opportunities}, the domains in which ML is deployed remain highly heterogeneous, requiring methods that respond to different constraints on data, computation, deployment, and interpretability \cite{Jordan2015trends}.
Within this heterogeneous landscape, a particularly important class of problems consists of bounded classification tasks, that must be solved under concrete operational constraints, e.g.: sensor-facing edge and IoT systems with limited memory, energy, bandwidth, and latency budgets \cite{Singh2023edge}, microcontroller-class on-device inference where cloud-based processing is impractical or undesirable \cite{Soroush2025tiny}, industrial fault detection and diagnosis under real-time conditions \cite{Leite2025fault}, and high-stakes decision-support settings where inherently interpretable models are preferred over post-hoc explanations of black boxes \cite{Rudin2019stop}. Taken together, such settings motivate classification architectures in which efficiency, deployability, and interpretability are not post hoc additions but core properties of the model itself.

In this context, Tsetlin Machines~\cite{granmo2018tsetlin} (TMs) have recently emerged as a compelling alternative to mainstream Machine Learning (ML) models, combining competitive accuracy with inherent explainability. At their core, TMs learn ensembles of propositional clauses over Booleanized input features: each clause is a conjunction of literals whose inclusion or exclusion is controlled by teams of Tsetlin automata, and classification is performed by aggregating the votes of clauses associated with each class~\cite{granmo2018tsetlin,Abeyrathna2024concise}. This representation makes the learned model inspectable as a set of logical patterns, while also enabling efficient execution through sparse clause evaluation, binary feature representations, and integer-valued voting mechanisms~\cite{Saha2023discover,Abeyrathna2024concise}. Recent hardware-oriented studies report that TM accelerators can support high-throughput and low-energy inference, including FPGA designs with on-device training and reported performance on image-classification benchmarks in the range of hundreds of thousands of classifications per second~\cite{Tunheim2025fpga,Mao2025dynamic}. This logic-based paradigm is particularly advantageous for custom hardware implementations, since the inference pipeline can be reduced largely to bitwise operations, clause evaluation, and additions, avoiding the floating-point multiplications and multiply--accumulate units that typically dominate the area, power, and thermal budget of conventional neural accelerators~\cite{Wheeldon2020automata,Tunheim2025fpga}. As a result, TMs are well aligned with deployment on resource-constrained and power-sensitive platforms, including edge devices, drones, autonomous robots, and satellites, particularly those tasked with carrying complex imaging sensors.

Among the most information-rich imaging modalities used in remote and embedded sensing is hyperspectral imaging (HSI). It acquires, for each spatial pixel, a spectral signature sampled across many narrow, typically contiguous, wavelength bands, forming a three-dimensional spatial--spectral data cube~\cite{Langer2023robust,Grewal2023survey}. This detailed spectral characterisation makes hyperspectral data highly valuable for sensor-based processing, with established applications in, e.g., remote sensing, environmental monitoring, precision agriculture, geological and mineral exploration, water-resource analysis, food-quality inspection, and medical analysis~\cite{Bhargava2024review,Ram2024agri,Wan2025hsi}. At the same time, HSI is a difficult computational target: the same spectral richness that enables fine material discrimination also produces high-dimensional, redundant, noisy, and nonlinear data, often combined with limited labelled samples, spectral variability, and mixed pixels~\cite{Grewal2023survey,Wan2025hsi}. These challenges are amplified in operational settings, where hyperspectral sensors may be mounted on satellites, drones, or other mobile platforms, and where massive data rates, limited onboard power, restricted memory, communication bottlenecks, and low-latency decision requirements make ground-based offline processing insufficient~\cite{Langer2023robust,Yang2025onboard}. Consequently, hyperspectral analysis has long motivated the development of dedicated fast-processing architectures, including FPGA, GPU, and heterogeneous FPGA--ARM designs, to meet near-real-time or onboard processing requirements~\cite{Gonzales2013fpga,Yang2025onboard}. For Tsetlin Machines, however, hyperspectral data also exposes a fundamental representational difficulty: spectra are continuous-valued signals, whereas standard TMs operate on propositional logic, Boolean literals, and bit-level inputs~\cite{granmo2018tsetlin}. Applying TMs to HSI therefore requires a careful interface between continuous spectral measurements and binary logical learning, where binarisation or adaptive continuous-feature representations must preserve discriminative spectral thresholds and inter-band structure without sacrificing the computational advantages of clause-based inference~\cite{abeyrathna2021adaptive}. A successful solution to this interface would be particularly valuable, as it would introduce a compact, interpretable, and hardware-friendly learning architecture into one of the most information-rich and operationally demanding domains of sensor-based machine learning.

To address this challenge, we propose a novel, high-performance binarisation algorithm inspired by the Viola-Jones rapid object detection framework~\cite{viola2001rapid}. Our algorithm introduces 1D Spectral Haar-like features that are highly efficient to compute. By utilising an integral spectrum, the feature extraction process—much like the underlying TM—relies predominantly on sequence additions, completely avoiding multiplications during inference. Furthermore, the algorithm is inherently explainable, as the resulting binary features map directly to straightforward spectral indices (differences between distinct wavelength intervals). This combination of computational efficiency and transparency makes the Spectral Haar Binarizer an ideal preprocessing counterpart to the Tsetlin Machine pipeline. Subsequently, we show that this binarisation algorithm outperforms all reference methods for both the TM and a baseline reference Random Forest classifier. Our findings show that the method outperforms the reference classifier trained on raw, unbinarised data. Importantly, the use of the proposed binarisation scheme is the only scenario in which TM is practically equivalent to a more complex RF classifier. Our results open the possibility of applying TM as an edge machine-learning solution for hyperspectral sensors.

\section{Spectral Haar Binarizer}\label{sec:method}
\subsection{Hyperspectral images}

A hyperspectral image can be represented as a three-dimensional data cube, with two spatial dimensions and one spectral dimension. Depending on the application, this cube may be analysed using spatial--spectral neighbourhoods, but many classification pipelines also operate at the level of individual spectra, treating each pixel as an independent measurement vector. This pixel-wise formulation is common in remote-sensing scenarios, where a distant airborne or satellite sensor records land-cover signatures over large areas~\cite{Melgani2004svm}, as well as in close-range forensic imaging, where hyperspectral cameras are used to detect and classify traces such as blood stains in crime-scene-like imagery~\cite{Romaszewski2021blood}. 

\begin{figure*}
    \centering
    \includegraphics[width=\textwidth]{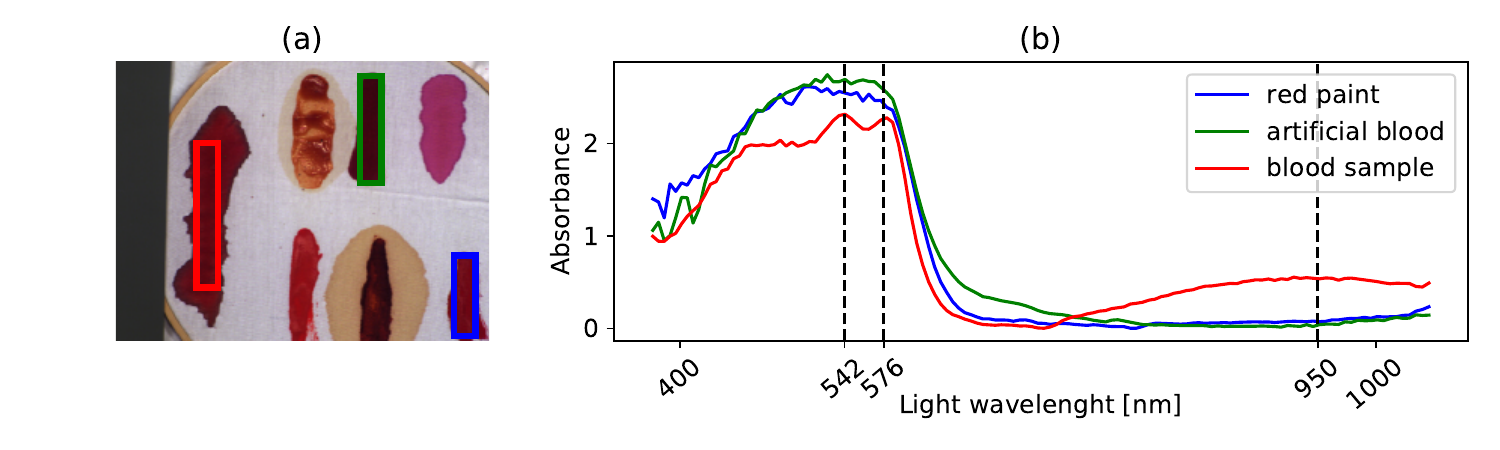}
    \caption{
    Hyperspectral imaging of blood and blood-like substances, based on the dataset introduced in~\cite{Romaszewski2021blood}.
    (a) Laboratory scene containing blood and visually similar red substances.
    (b) Absorbance spectra of blood, red paint, and artificial blood.
    Although these materials may appear similar in RGB imagery, their hyperspectral signatures reveal distinct spectral characteristics.
    In particular, the blood spectrum exhibits characteristic haemoglobin absorption features, including the $\alpha$ and $\beta$ bands around 576~nm and 542~nm, respectively, as well as a further feature near 950~nm.
    This illustrates the potential of hyperspectral imaging for discriminating substances and analysing their material properties beyond what is accessible from conventional RGB data.
    }
    \label{fig:blood}
\end{figure*}

Let $\mathbf{x} = [x_1, x_2, \dots, x_B]^T$ denote such a spectral signal, corresponding, for example, to a single hyperspectral pixel, where $B$ is the number of acquired spectral bands. Each entry $x_b$ describes the measured response at one wavelength band, so that $\mathbf{x}$ represents one complete spectral profile extracted from the hyperspectral cube. In terms of Fig.~\ref{fig:blood}, a single vector $\mathbf{x}$ corresponds to one of the plotted absorbance curves in panel~(b), rather than to the full image shown in panel~(a). This pixel-wise representation is the input to the classifier.

\subsection{Binarisation for Tsetlin Machines}

Tsetlin Machines operate on Boolean input variables; consequently, applying them to continuous spectral measurements requires an explicit mapping from the original feature space to a binary representation. Therefore, a spectral vector $\mathbf{x} \in \mathbb{R}^{B}$ is mapped to the Boolean vector $\mathbf{z}$
\[
    \Phi: \mathbb{R}^{B} \rightarrow \{0,1\}^{D},
    \qquad \mathbf{z} = \Phi(\mathbf{x}).
\]
This formulation follows directly from the original TM model, in which learning and inference are defined over propositional variables and their negations~\cite{granmo2018tsetlin}. For continuous-valued inputs, previous TM work has therefore introduced threshold-based encodings that convert each scalar feature into a set of Boolean tests~\cite{abeyrathna2019continuous}. In the present formulation, this idea is expressed by first assigning each band value $x_b$ to one of $K_b$ discrete bins $q_b(x_b) \in \{0,1,\dots,K_b-1\}$, and then encoding the resulting bin index into Boolean variables.

Two encodings are considered. In the thermometer variant, the bin index is represented cumulatively:
\[
    z^{\mathrm{TH}}_{b,j} =
    \mathbb{I}\left[q_b(x_b)\geq j\right],
    \qquad j=1,\dots,K_b-1.
\]
This encoding preserves the ordinal structure of the discretised feature and is closely related to earlier threshold-based continuous-input schemes for TMs, where Boolean variables represent ordered tests such as $x_b \geq \tau_j$~\cite{abeyrathna2019continuous}. In this sense, thermometer encoding follows the traditional TM approach to continuous inputs: it exposes threshold-exceedance predicates that can be directly combined by clauses into logical rules.

In the one-hot variant, the same bin index is instead represented as a Boolean vector whose $j$-th component is active only when the value of band $b$ falls into bin $j$:
\[
    z^{\mathrm{OH}}_{b,j} =
    \mathbb{I}\left[q_b(x_b)=j\right],
    \qquad j=0,\dots,K_b-1.
\]
This encoding extends the threshold-oriented TM binarisation scheme toward interval-membership predicates. After discretisation, each bin can be treated as a categorical state of the spectral feature, and one-hot encoding is a standard way to represent such categorical values as binary variables~\cite{Garcia2013survey,Hancock2020categorical}. Thus, while thermometer encoding exposes ordered threshold-exceedance conditions, one-hot encoding exposes localised value-range conditions; both define the Boolean predicates that TM clauses may combine into logical rules.

The choice of binarisation is particularly important for hyperspectral data. Spectral measurements are continuous-valued and often contain subtle absorption features distributed across neighbouring bands, whereas standard TMs learn clauses over Boolean literals. The binarisation procedure defines the basic Boolean conditions that TM clauses can combine to form logical rules and, consequently, the spectral thresholds, intervals, or value ranges that the model can express. If the discretisation is too coarse, discriminative spectral information may be lost; if it is too dense, the Boolean dimensionality increases and clause learning may become more difficult. The goal of the binarisation step is therefore to preserve class-relevant spectral structure while maintaining the compact Boolean representation required for efficient TM inference.

\subsection{Proposed 1D Haar-like Spectral Features}

To process (hyper)spectral data for bit-based models such as Tsetlin Machines, we propose a 1D Spectral Haar Binariser. The proposed binarisation scheme is inspired by the rectangular contrast features introduced in the Viola--Jones face detector~\cite{Viola2004robust}. In that framework, simple Haar-like features compare the sums of pixel intensities across adjacent rectangular regions, allowing local contrast patterns to be captured efficiently rather than relying solely on raw pixel values. Their effectiveness comes from combining simple, interpretable feature responses with efficient computation through integral images and feature selection. We adapt this principle to spectral data by constructing binary predicates from contrasts between selected regions of the spectral signal.

\subsubsection{Feature Extraction}

The algorithm extracts a compact set of binary features by measuring the normalised difference in spectral intensities between pairs of intervals. We define a 1D Haar-like feature $f(\mathbf{x})$ based on a pair of spectral intervals $I_1 = [a_1, c_1]$ and $I_2 = [a_2, c_2]$, where $1 \le a_i \le c_i \le B$. The feature response is calculated as the difference between the mean intensity within these two intervals:
\begin{equation}
    f(\mathbf{x}) = \frac{1}{L_1} \sum_{k=a_1}^{c_1} x_k - \frac{1}{L_2} \sum_{k=a_2}^{c_2} x_k
\end{equation}
where $L_1 = c_1 - a_1 + 1$ and $L_2 = c_2 - a_2 + 1$ are the lengths of the respective intervals. Normalising by the interval length accounts for differences in scale, rendering the features comparable regardless of their spectral bandwidth.

To compute the interval sums efficiently, we employ a 1D integral spectrum (prefix sum) analogous to the 2D integral image used by Viola and Jones. Let $P$ be the integral spectrum defined as $P_k = \sum_{t=1}^{k} x_t$ with $P_0 = 0$. The sum of any interval $[a, c]$ can then be evaluated in constant $\mathcal{O}(1)$ time:
\begin{equation}
    \sum_{k=a}^{c} x_k = P_c - P_{a-1}
\end{equation}

\subsubsection{Feature Binarisation}

Since our target model requires a binary input space $\mathbf{z} \in \{0, 1\}^K$, we threshold each continuous feature value. Analogous to decision stumps, each threshold test defines a simple binary predicate over a continuous feature, indicating whether the observed value exceeds a selected threshold. For the $j$-th feature, the binary activation $z_j$ is given by:
\begin{equation}
    z_j = 
    \begin{cases} 
        1 & \text{if } s_j (f_j(\mathbf{x}) - \theta_j) \ge 0 \\
        0 & \text{otherwise}
    \end{cases}
\end{equation}
where $\theta_j$ is a scalar threshold and $s_j \in \{-1, +1\}$ denotes the polarity of the inequality. 

\subsubsection{Feature Selection}

Because the space of all possible interval pairs $(I_1,I_2)$ is large, we first generate a candidate pool of $M$ randomly sampled interval-pair features. The sampling procedure includes three types of spectral configurations: adjacent, contained, and partially overlapping intervals. These patterns are intended to capture different forms of spectral contrast, including local changes, band-specific deviations, and broader relationships between neighbouring spectral regions. To keep the candidates local and computationally manageable, interval lengths and relative positions are constrained by a maximum window size.

From the candidate pool, we select a compact subset of $K$ features, where $K \ll M$, using an AdaBoost-inspired procedure. The purpose of this step is not to train the final classifier, but to select a small set of interval-pair contrasts that separate the classes well while progressively focusing on samples that remain difficult to classify.

Following the Viola--Jones methodology, each candidate contrast feature is converted into a set of simple threshold tests. Analogous to decision stumps, each threshold test defines a binary predicate over a continuous feature value, indicating whether the observed contrast exceeds a selected threshold. For a candidate feature $f_m$, threshold $\theta$, and polarity $p \in \{-1,+1\}$, the corresponding stump can be written as
\[
    h_{m,\theta,p}(\mathbf{x}) =
    \begin{cases}
    +1, & p f_m(\mathbf{x}) \geq p\theta,\\
    -1, & \text{otherwise}.
    \end{cases}
\]

At boosting round $t$, each training sample $i$ is assigned a weight $w_i^{(t)}$. All candidate stumps are evaluated by testing multiple thresholds and both polarities for each candidate feature. The stump with the smallest weighted classification error is selected:
\[
    \epsilon_t =
    \sum_{i : h_t(\mathbf{x}_i) \neq y_i}
    w_i^{(t)} .
\]
The interval-pair feature associated with the selected stump is added to the final feature set, together with the selected threshold and polarity. The sample weights are then updated according to the standard AdaBoost rule,
\[
    w_i^{(t+1)}
    =
    w_i^{(t)}
    \exp\left(
        -\alpha_t y_i h_t(\mathbf{x}_i)
    \right),
\]
where
\[
    \alpha_t =
    \frac{1}{2}
    \ln
    \left(
        \frac{1-\epsilon_t}{\epsilon_t}
    \right).
\]
This reweighting increases the importance of samples misclassified by the current stump, encouraging subsequent rounds to select features that capture complementary discriminative information.

For multiclass problems, we use a one-vs-rest strategy. In each class-specific selection problem, samples from the target class are treated as positive, and all remaining samples are treated as negative. Boosting rounds are distributed across classes so that the selected feature set contains predicates discriminative for each target class. The final output of the selection stage is therefore a compact collection of spectral contrast predicates, which are subsequently used as Boolean inputs for the Tsetlin Machine.

\subsubsection{Implementation and Computational Optimisation}
A defining characteristic of Tsetlin Machines is their reliance on propositional logic rather than floating-point arithmetic, making them extremely efficient during inference. The Viola-Jones framework shares a similar philosophy: by utilising the integral image, feature extraction at inference operates entirely through integer additions and subtractions, avoiding costly multiplications and divisions. 

Our 1D Spectral Haar Binarizer adapts this principle via the integral spectrum. However, because the selected interval pairs natively vary in length ($L_1$ and $L_2$), direct comparison of their sums would be unbalanced. In the training phase, the algorithm calculates the feature response as the difference of interval means:
\begin{equation}
    \frac{1}{L_1} S_1 - \frac{1}{L_2} S_2 \ge \theta
\end{equation}
where $S_1 = \sum_{k=a_1}^{b_1} x_k$ and $S_2 = \sum_{k=a_2}^{b_2} x_k$ are the interval sums extracted in constant time. 

While evaluating means entails floating-point division, the process can be strictly optimised for inference. By algebraically rearranging the inequality to eliminate the denominators, the inference condition becomes:
\begin{equation}
    L_2 S_1 - L_1 S_2 \ge \theta L_1 L_2
\end{equation}
Because $L_1$, $L_2$, and the threshold $\theta$ are fixed constants determined during training~\footnote{e.g. $\theta$ values are selected from a fixed vector of 10 percentiles from 5 to 95 for each feature, instead of all midpoint candidates between sorted feature values}, the term $\theta L_1 L_2$ can be precomputed as a single scaled threshold, $\theta_{scaled} = \theta L_1 L_2$. At inference time, calculating a feature's activation simply requires evaluating the boolean statement:
\begin{equation}
    L_2 S_1 - L_1 S_2 \ge \theta_{scaled}, 
\end{equation}

With this optimisation, when spectral values are quantised to integers, the entire inference pipeline operates in integer arithmetic. While the proposed optimisation still requires integer multiplication by a known constant (scaling the respective sums by the opposing interval's length), on hardware this can be implemented as shifts and additions.

We again note that the implementation employs AdaBoost-style processing strictly as a feature selection mechanism. In traditional applications, AdaBoost generates a final prediction by calculating a weighted sum of weak classifiers using floating-point weights ($\alpha_t$). In contrast, our pipeline deliberately discards these weights after the salient features are identified, evaluating only the unweighted Boolean outputs ($z_j \in \{0, 1\}$). 

We also omit the polarity check from this discussion, as it reduces to a single conditional negation.

\subsection{Reference Methods}

To evaluate the proposed binarisation strategy, we compare it against four reference binarizers that instantiate different principles for constructing the mapping $\Phi$ from continuous spectral values to Boolean variables. The Standard Binarizer represents the threshold-based encoding commonly used with Tsetlin Machines for continuous inputs~\cite{granmo2018tsetlin,abeyrathna2019continuous}. The Multi-Otsu Binarizer provides a histogram-based multilevel thresholding baseline, motivated by prior TM applications in which non-binary image or spectrogram representations are converted to Boolean form before clause learning~\cite{granmo2019convolutional,jeeru2025mixctme}. The Quantile Binarizer provides a distribution-balanced discretisation baseline, in which thresholds are chosen so that the bins contain approximately equal numbers of samples. The K-Means Binarizer provides a complementary quantisation-based baseline, in which continuous observations are represented by the nearest element in a finite set of prototype values~\cite{Lloyd1982least}. Together, these methods cover monotonic threshold encoding, variance-based multilevel thresholding, quantile-based binning, and clustering-based quantisation.

In addition to the Tsetlin Machine classifier, we use a Random Forest (RF) classifier as a reference. RF is a well-established ensemble of randomised decision trees~\cite{Breiman2001random} and has been widely used as a strong non-neural baseline in remote-sensing and hyperspectral classification~\cite{Pal2005random,Joelsson2005random}. It provides a useful comparison point because, like TMs, it is based on combinations of simple decision rules rather than dense continuous representations, and it does not require Boolean input features.

\subsubsection{Standard Binarizer}

The Standard Binarizer discretises each continuous feature independently into a sequence of Boolean variables using thresholding. This corresponds to the traditional continuous-input TM strategy, in which scalar values are converted into ordered threshold tests before clause learning~\cite{abeyrathna2019continuous}. For a given feature, the implementation determines the sorted set of unique values. If the number of distinct values exceeds the predefined hyperparameter $n_{\mathrm{bit}}$, at most $n_{\mathrm{bit}}$ thresholds are selected from the ordered empirical values. A Boolean indicator is then generated for each threshold, evaluating to $1$ if the observed feature value is greater than or equal to the threshold and $0$ otherwise. This produces a monotonic, thermometer-like representation in which larger feature values activate a nested set of Boolean variables. We use the implementation provided by the official TM library \texttt{pyTsetlinMachine}\footnote{\url{https://github.com/cair/pyTsetlinMachine}}.

\subsubsection{Multi-Otsu Binarizer}

The Multi-Otsu Binarizer adapts histogram-based thresholding to the one-dimensional distribution of each spectral feature. Otsu's method was originally introduced for automatic threshold selection in image segmentation, where thresholds are chosen to maximise the separability of grey-level classes~\cite{Otsu1979threshold}. Its multilevel extension generalises this idea to multiple thresholds, producing $k$ discrete classes from $k-1$ thresholds~\cite{Liao2001fast}. In the present setting, Multi-Otsu is applied independently to each spectral band, searching for thresholds that maximise the separation between value groups in the empirical feature histogram. The resulting bin index is then encoded as a Boolean vector using either one-hot or thermometer encoding.

This baseline is motivated by prior TM applications in image-like data, where non-binary sensor measurements are thresholded before clause learning. For example, the Convolutional Tsetlin Machine uses binary image layers and applies adaptive Gaussian thresholding to grayscale inputs~\cite{granmo2019convolutional}, while MixCTME uses Otsu-binarised spectrogram visualisations for convolutional TM experts~\cite{jeeru2025mixctme}. Although these works operate on image or spectrogram representations rather than hyperspectral pixel spectra, they provide a relevant precedent for using thresholding-based transformations to expose structured sensor data to Boolean TM clauses. The thresholding operation is implemented using the \texttt{threshold\_multiotsu} function from \texttt{scikit-image}\footnote{\url{https://scikit-image.org/docs/stable/api/skimage.filters.html\#skimage.filters.threshold_multiotsu}}.

\subsubsection{Quantile Binarizer}

The Quantile Binarizer discretises each continuous spectral feature by placing thresholds at empirical percentiles of its distribution. For $k$ bins, the bin edges are defined by the percentiles
\[
    \frac{100}{k}, \frac{2 \cdot 100}{k}, \dots, \frac{(k-1)\cdot 100}{k}.
\]
For example, when $k=5$, thresholds are placed at the 20th, 40th, 60th, and 80th percentiles; when $k=2$, the threshold is the median. Each observation is assigned to the corresponding quantile bin, and the resulting bin index is encoded as a Boolean vector using either one-hot or thermometer encoding.

This baseline complements the Standard, Multi-Otsu, and K-Means binarizers by enforcing approximately balanced bin occupancy. Rather than using fixed empirical thresholds, maximising histogram separability, or fitting representative centroids, quantile binning adapts to the marginal distribution of each spectral band while avoiding bins that contain very few samples. This is particularly relevant for skewed or uneven feature distributions, where uniformly placed thresholds may allocate most observations to a small number of bins.

\subsubsection{K-Means Binarizer}

The K-Means Binarizer provides a quantisation-based discretisation strategy for continuous spectral features. It is motivated by Lloyd-style least-squares quantisation, where continuous observations are represented by the nearest element of a finite set of prototype values~\cite{Lloyd1982least}. In the present setting, this idea is applied independently to each spectral band by performing one-dimensional K-Means clustering on its empirical distribution. The resulting centroids define $k$ representative value regions, and each observation is assigned to the bin corresponding to the nearest centroid.

The bin assignment is then transformed into a Boolean representation. In the one-hot variant, each bin corresponds to a separate Boolean variable, allowing TM clauses to test whether a spectral value falls into a particular data-driven region. In the thermometer variant, the same bin index is encoded cumulatively, preserving the ordering of the centroid-induced bins. This tests whether TM clauses benefit from spectral bins that follow the observed data distribution, for example, when feature distributions are uneven, skewed, or clustered, rather than from fixed threshold placement or variance-based threshold separation. Compared with fixed or empirical threshold placement, this baseline therefore evaluates whether adapting bin boundaries to the distribution of each spectral band improves the Boolean predicates available to the TM. The implementation is based on \texttt{KBinsDiscretizer} from \texttt{scikit-learn}, using the K-Means discretisation strategy\footnote{\url{https://scikit-learn.org/stable/modules/generated/sklearn.preprocessing.KBinsDiscretizer.html}}.

\section{Experimental Validation of the Proposed Method}

\subsection{Datasets}

We evaluate the proposed approach on a deliberately heterogeneous set of five hyperspectral datasets: Indian Pines (INPI), Pavia University (PAVI), La Selva mosaic (MOSA), and two scenes from the hyperspectral blood detection dataset (HBE1 and HBF1). This selection is intended to test the method beyond a single sensor, scene type, or application domain. The benchmark includes airborne agricultural and forest remote sensing, urban material classification, tropical tree-species discrimination, and close-range forensic substance detection. It also spans different spatial resolutions, spectral ranges, numbers of bands and classes, and acquisition conditions. As a result, the evaluation probes whether the proposed binarisation strategy is robust not only to standard hyperspectral classification benchmarks, but also to substantially different spectral signatures, class semantics, and operational settings.

\paragraph{Indian Pines (INPI)}
This hyperspectral image was collected by the AVIRIS sensor over the Indian Pines region in Northwestern Indiana in 1992, covering a mixed agricultural and forest area early in the growing season. The image possesses a spatial resolution of 20 m per pixel. Following the removal of channels affected by noise and water absorption (104–108, 150–163, and 220), 200 spectral channels remain. The ground reference data contains 16 distinct classes, representing different types of crops.

\paragraph{University of Pavia (PAVI)}
The Pavia image was captured using the ROSIS optical sensor over the urban area of the University of Pavia, Italy. After discarding regions lacking class information, the valid image size features 115 spectral channels (with 12 noisy bands) spanning the 0.43–0.86 \textmu m spectral range, with a spatial resolution of 1.3 m per pixel. The ground reference data delineates 9 classes representing diverse urban materials, including gravel, trees, metal, bare soil, bricks, and shadows.

\paragraph{La Selva Biological Station (MOSA)}
This dataset was captured by the HYDICE hyperspectral sensor over the La Selva Biological Station in March 1998 and was created to test methods for automated species-level classification of individual tree crowns in a tropical rainforest. The processing and georeferencing procedures are detailed in~\cite{clark2005hyperspectral}. The original image was registered from multiple runs and contains atmospherically corrected surface reflectance of tree canopies. It features 161 spectral bands spanning 437.2 to 2434.3 nm, with a high spatial resolution of 1.6 m per pixel. The dataset contains 7 classes representing specific tree species. Our experiments are performed on a fragment of a mosaic composed of multiple passes, in which samples from all classes are present.

\paragraph{Blood Hyperspectral Dataset (HBE1, HBF1)}
The Blood Hyperspectral dataset consists of hyperspectral images acquired from a mock forensic scene containing real blood, artificial blood, and visually similar substances such as tomato concentrate, ketchup, beetroot juice, and red paints. Acquisitions were performed with a Surface Optics SOC710 VNIR hyperspectral camera covering the 377–1046 nm range, producing images of $696 \times 520$ pixels with 128 spectral bands and 12-bit radiometric resolution~\cite{Romaszewski2021blood}. After removing noisy bands, following the dataset creators' suggestions, 113 bands remain.
We use two scenes from the dataset:
\begin{itemize}
    \item \textbf{Scene E (HBE1 - `Comparison'):} The scene tests classifiers in a challenging, diverse environment. It contains blood and five blood-like substances arranged across eight different backgrounds (fabrics, wood, plastic, and metal), some of which feature a red hue similar to the detected targets.
    \item \textbf{Scene F (HBF1 - `Frame'):} Is intended as a controlled source of blood spectra; this scene contains blood and six blood-like substances on a piece of white fabric stretched on a wooden frame.
\end{itemize}

\subsubsection{Classification Setup} 

Because both the proposed binarisation method and standard Tsetlin Machines operate on binary inputs, we construct binary classification tasks by selecting pairs of classes from each dataset, and we also evaluate the standard multiclass setting.

\paragraph{Indian Pines (INPI)}
In the INPI dataset, class pairs are selected to represent different types of agricultural classification difficulty. The pair \textit{Corn-notill} vs \textit{Soybean-notill} explores spectral similarity between different crop types. The pair \textit{Soybean-mintill} vs \textit{Corn-mintill} examines discrimination between crops under similar tillage conditions. The pair \textit{Woods} vs \textit{Grass-pasture} represents a difference between broader vegetation types.

\paragraph{Pavia University (PAVI)}
In the PAVI dataset, class pairs are chosen based on similarity in material or vegetation characteristics. The pair \textit{Asphalt} vs \textit{Bitumen} represents materials with similar spectral properties. The pair \textit{Meadows} vs \textit{Trees} reflects differences between low vegetation and tree cover.

\paragraph{La Selva Biological Station (MOSA)}
In the MOSA dataset, the class pairs \textit{BAEL} vs \textit{DIPA} and \textit{LEAM} vs \textit{CEPE} were selected because they share a similar phenological pattern (deciduous species with annual leaf exchange) but differ in their expected leaf cover at the time of image acquisition (end of dry season).

\paragraph{Blood Datasets (HBE1, HBF1)}
For the hyperblood dataset, the binary task (\textit{Blood} vs \textit{Non-Blood}) aligns with the main objective of~\cite{Romaszewski2021blood}, which evaluates the detection of blood in the presence of visually similar substances.

\subsection{Experimental protocol}

For each dataset, we define a set of classification tasks, including selected binary classification problems and the standard multiclass classification setting. For each dataset--task pair, we evaluate every combination of binarisation method and classifier, including the proposed binarisation approach and the reference binarisation methods paired with both the Tsetlin Machine (TM) and Random Forest (RF) classifiers. For multiclass problems, we use multiclass TM and RF classifiers as described in Sec.~\ref{sec:method}, and additionally include a canonical RF baseline trained directly on the unbinarized raw spectra.

Each such configuration is evaluated using Monte Carlo cross-validation (MCCV). Specifically, we generate five independent random splits of the data, each using a different random seed. In every split, the samples are partitioned into 50\% training, 25\% validation, and 25\% testing subsets. Binarisation parameters, model and hyperparameter selection are performed within each split using the training and validation subsets, while the test subset is held out and used only for the final evaluation of the selected configuration. The resulting predictions are assessed using the metrics described in the Evaluation section.

\subsubsection{Hyperparameter Selection}

Classifier hyperparameters are selected independently for each experimental configuration using internal cross-validation on the training data. This is particularly important for Tsetlin Machines, whose performance depends on parameters such as the number of clauses, the voting threshold, and the specificity parameter $s$; the difficulty of selecting suitable TM parameter granularities has been explicitly discussed in prior work on multigranular clauses~\cite{rahimi2019tsetlin}. Using the same selection protocol across all binarisation methods ensures that the comparison reflects the effect of the binarisation strategy rather than manually optimised TM settings. For the Random Forest (RF) classifier, we evaluate maximum tree depths of $\{\texttt{None}, 5, 10\}$; remaining hyperparameters are set to default recommended values. For the Tsetlin Machine (TM), we evaluate combinations of the number of clauses $\in \{80, 120, 180, 250, 300\}$, specificity $s \in \{2.0, 2.5, 3.0, 3.5, 4.0, 7.5\}$, voting thresholds $\in \{12, 14, 15, 22, 30, 35\}$, and training epochs $\in \{40, 50, 60, 70, 100\}$.

The proposed Spectral Haar Binarizer (SHB) is configured to select $256$ features from an initial candidate pool of $1024$, with a maximum window size of $25$ contiguous spectral bands. For each train/validation split, the optimal classifier parameter combination is selected based on the highest validation Balanced Accuracy, and the final classification metrics are reported on the separate test set.

For the reference binarizers, the number of output bits/bins per spectral feature is fixed to $5$ for all datasets and classification tasks. We treat this value as part of the input representation rather than as a classifier hyperparameter optimised separately for each run. This follows the common use of TM pipelines, where the Boolean representation is either assumed by the method, imposed by the data representation, or fixed as a preprocessing design choice, while the classifier parameters are then selected on top of this representation.

The value of $5$ bits/bins was chosen as a compromise between quantisation precision and the dimensionality of the resulting Boolean feature space. This choice was guided by two considerations: first, the scale of binarisation settings used in related TM work on continuous inputs, and second, preliminary experiments comparing classification performance with representation size. Both indicated that $5$ bits/bins provide a practical balance: lower values tend to lose spectral detail, whereas higher values increase the Boolean dimensionality and computational cost without consistently improving validation performance. Since the number of bins is nevertheless an important design parameter, we analyse its influence separately in Sec.~\ref{sec:ablation_bits}.

\subsubsection{Evaluation}

We evaluate classification performance using standard predictive metrics: Accuracy, Balanced Accuracy, and F1-score. Accuracy represents the overall fraction of correctly classified samples, while Balanced Accuracy and F1-score are particularly important in our setting, as several datasets and binary tasks exhibit class imbalance. Balanced Accuracy is used as the primary metric for model selection and method comparison because it gives equal weight to all classes by averaging per-class recall.

Although these metrics provide a direct view of predictive performance, they are not sufficient on their own to determine whether one method is consistently better than another. In particular, comparing mean scores and standard deviations across many datasets, class pairs, binarisation schemes, and classifiers can be difficult to interpret, especially when the observed differences are small. Therefore, we complement the classical metrics with two aggregate analyses. First, we use Bayesian statistical comparison to estimate whether the proposed method is practically better, worse, or equivalent to the reference methods. Second, we use a rank-based scoring table to summarise the distribution of strong performances across the full set of experiments.

\paragraph{Bayesian Comparison}
To evaluate the relative performance of the proposed Spectral Haar Binariser (SHB) against the established baseline discretisation methods, we employ a Bayesian statistical framework~\cite{benavoli2017time}. Unlike frequentist null-hypothesis significance testing (NHST), Bayesian analysis provides direct probability estimates that one classifier is practically better, worse, or equivalent to another. 

We utilise the Bayesian Hierarchical Correlated $t$-test~\cite{benavoli2017time}, which is designed to compare classifiers across multiple datasets evaluated via cross-validation. This hierarchical model jointly estimates the intra-dataset variance, which reflects noise across different folds, and the inter-dataset variance, which reflects differences in task difficulty across datasets.

We evaluate the binary and multiclass domains independently. To account for the fact that small differences in performance are often practically meaningless, the framework in~\cite{benavoli2017time} defines a Region of Practical Equivalence (ROPE). We set this value to $1\%$ ($\tau = 0.01$), implying that two methods whose average Balanced Accuracy scores differ by less than $1$ percentage point are considered practically equivalent. The hierarchical test yields a posterior distribution over the performance differences, which we integrate to compute three probabilities: the probability that the control method is better ($P(C > B)$), the probability that the baseline is better ($P(B > C)$), and the probability that the two methods are practically equivalent ($P(C \approx B)$).

\paragraph{Rank-based Scoring}\label{sec:results_scoring}
As an additional descriptive summary, we employ a rank-based scoring system. This analysis is not intended to replace the Bayesian comparison; rather, it provides a compact overview of which methods repeatedly appear among the best-performing configurations across many datasets and classification tasks. Such a summary is useful because the experimental design yields many individual results, and small differences in average performance can obscure whether a method is broadly reliable or succeeds only on a small subset of tasks.

For each individual experiment, defined by a dataset and, for binary problems, a class pair, all classifier--binarizer combinations are ranked by a chosen test-set metric, Balanced Accuracy by default. Points are awarded according to placement: 3~points for first place, 2~points for second place, and 1~point for third place; all remaining methods receive zero. To avoid overinterpreting negligible score differences, the same tolerance parameter used in the Bayesian analysis is applied when assigning ranks. Specifically, all methods whose metric value falls within the ROPE threshold $\tau = 0.01$ of the current rank group's leader are assigned the same rank.

This scoring scheme has three practical advantages. First, it treats each dataset--task instance as one unit of evidence, preventing large or easy tasks from dominating the summary solely through sample count. Second, the ROPE-aware ranking avoids rewarding methods for differences that are too small to be practically meaningful. Third, by awarding points to the top three rank groups rather than only to the winner, it captures both outright best performance and repeated near-best performance. The resulting point totals are accumulated across all experiments and presented in a summary table. Rows correspond to individual experiments, grouped into binary and multiclass sections, columns correspond to classifier--binarizer combinations, and a final row reports the aggregate score for each method.

\subsection{Results}

\subsubsection{Standard Metrics}

\begin{table*}
\centering
\caption{Averaged performance of binarizer-classifier pairs across all datasets. Shortcuts: B: Binary, M: Multiclass, MO: MultiOtsu, T: Thermometer, OH: OneHot, Quant: Quantile. Methods ordered by F1.}
\begin{tabular}{llrrr}
\toprule
Task& Method & Avg Acc & Avg BAcc & Avg F1 \\
\midrule
\multirow{17}{*}{B} & TM-SHB & $97.08 \pm 0.29$ & $94.89 \pm 0.58$ & $95.58 \pm 0.43$ \\
 & RF-SHB & $96.81 \pm 0.41$ & $94.50 \pm 0.73$ & $95.31 \pm 0.61$ \\
 & RF-Raw & $94.28 \pm 0.58$ & $90.50 \pm 0.94$ & $91.65 \pm 0.82$ \\
 & RF-Std & $92.95 \pm 0.72$ & $88.11 \pm 1.17$ & $89.26 \pm 1.01$ \\
 & RF-MO-T & $92.92 \pm 0.65$ & $87.77 \pm 1.18$ & $89.02 \pm 0.93$ \\
 & RF-MO-OH & $92.58 \pm 0.55$ & $87.22 \pm 1.11$ & $88.55 \pm 0.81$ \\
 & RF-KM-T & $92.50 \pm 0.59$ & $87.15 \pm 1.19$ & $88.35 \pm 0.92$ \\
 & RF-Quant-T & $92.29 \pm 0.60$ & $87.03 \pm 1.12$ & $88.20 \pm 0.90$ \\
 & RF-KM-OH & $92.20 \pm 0.83$ & $86.70 \pm 1.39$ & $87.99 \pm 1.21$ \\
 & RF-Quant-OH & $91.91 \pm 0.71$ & $86.28 \pm 1.36$ & $87.61 \pm 1.10$ \\
 & TM-MO-OH & $92.03 \pm 0.72$ & $85.60 \pm 1.41$ & $87.13 \pm 1.10$ \\
 & TM-Quant-OH & $91.79 \pm 0.78$ & $85.15 \pm 1.72$ & $86.64 \pm 1.41$ \\
 & TM-Quant-T & $91.01 \pm 1.19$ & $86.65 \pm 1.30$ & $86.28 \pm 1.43$ \\
 & TM-KM-OH & $91.57 \pm 0.93$ & $84.32 \pm 1.98$ & $86.09 \pm 1.56$ \\
 & TM-Std & $91.68 \pm 0.59$ & $83.62 \pm 1.72$ & $85.57 \pm 1.57$ \\
 & TM-MO-T & $91.14 \pm 0.82$ & $83.88 \pm 1.39$ & $85.15 \pm 1.30$ \\
 & TM-KM-T & $90.96 \pm 0.96$ & $83.66 \pm 1.58$ & $84.90 \pm 1.48$ \\
\midrule
\multirow{17}{*}{M} & RF-SHB & $93.11 \pm 0.38$ & $89.48 \pm 1.14$ & $90.75 \pm 0.83$ \\
 & RF-Raw & $89.84 \pm 0.33$ & $84.77 \pm 0.61$ & $86.41 \pm 0.57$ \\
 & TM-SHB & $90.93 \pm 0.68$ & $85.11 \pm 1.23$ & $86.02 \pm 2.89$ \\
 & RF-MO-OH & $84.05 \pm 0.31$ & $77.92 \pm 0.98$ & $79.41 \pm 0.96$ \\
 & RF-MO-T & $83.99 \pm 0.37$ & $77.86 \pm 1.06$ & $79.39 \pm 0.93$ \\
 & RF-Quant-T & $84.27 \pm 0.29$ & $77.40 \pm 0.51$ & $78.90 \pm 0.49$ \\
 & RF-Std & $83.93 \pm 0.41$ & $77.32 \pm 0.68$ & $78.73 \pm 0.64$ \\
 & RF-Quant-OH & $84.11 \pm 0.34$ & $77.20 \pm 0.55$ & $78.69 \pm 0.58$ \\
 & RF-KM-OH & $83.37 \pm 0.49$ & $76.82 \pm 0.93$ & $78.53 \pm 0.85$ \\
 & RF-KM-T & $83.28 \pm 0.41$ & $76.71 \pm 0.81$ & $78.39 \pm 0.75$ \\
 & TM-MO-OH & $80.37 \pm 0.54$ & $71.54 \pm 1.23$ & $72.50 \pm 1.72$ \\
 & TM-KM-OH & $79.80 \pm 0.58$ & $70.94 \pm 0.85$ & $71.31 \pm 2.48$ \\
 & TM-Quant-OH & $80.60 \pm 0.71$ & $71.56 \pm 0.97$ & $70.92 \pm 2.57$ \\
 & TM-Quant-T & $80.02 \pm 0.92$ & $70.67 \pm 0.99$ & $70.61 \pm 2.63$ \\
 & TM-KM-T & $78.71 \pm 0.62$ & $69.92 \pm 0.86$ & $70.61 \pm 1.35$ \\
 & TM-MO-T & $79.23 \pm 0.78$ & $70.90 \pm 1.07$ & $70.33 \pm 1.66$ \\
 & TM-Std & $78.86 \pm 0.48$ & $69.64 \pm 1.03$ & $69.97 \pm 2.02$ \\
\bottomrule
\end{tabular}
\label{tab:performance}
\end{table*}

Table~\ref{tab:performance} summarises the average predictive performance of all classifier--binarizer combinations. The proposed Spectral Haar Binarizer (SHB) achieves the strongest results in both evaluation regimes. For binary classification tasks, TM-SHB achieves the best overall performance, with an average Balanced Accuracy of $94.89\%$, followed closely by RF-SHB at $94.50\%$. For multiclass tasks, RF-SHB obtains the best performance, reaching $89.48\%$ Balanced Accuracy, while TM-SHB remains the strongest TM-based configuration and also outperforms the RF classifier trained directly on raw spectra in terms of Accuracy and Balanced Accuracy. These results indicate that SHB provides a substantially more effective Boolean representation than the reference binarisation schemes, especially for Tsetlin Machines, where the quality of the binary input representation directly determines the predicates available to the clauses.

The comparison with RF-Raw is particularly informative. Since RF-Raw operates on the original continuous spectra, it represents a strong non-binarised baseline. Nevertheless, SHB improves over  RF-Raw in both binary and multiclass settings, suggesting that the proposed spectral contrast features not only preserve information during binarisation but also expose discriminative structure useful for classification. In contrast, the conventional binarisation baselines generally perform worse than RF-Raw, especially in the multiclass setting. This pattern highlights the central role of binarisation design: reference encodings provide TM-compatible inputs, but they do not capture spectral structure as effectively as the proposed SHB representation. This pattern motivates the more detailed statistical comparison in the following sections, where we examine whether the observed differences persist across tasks and datasets.

\subsubsection{Individual Dataset Results}

The HBE1 results in Tab.~\ref{tab:results_hbe1} follow the aggregate trend from Tab.~\ref{tab:performance}. SHB-based methods rank first in both settings: RF-SHB leads the multiclass task, while TM-SHB performs best in the Blood-vs-Non-Blood task. This is particularly relevant for the forensic scenario, where blood must be separated from visually similar red substances; in this setting, SHB improves over RF-Raw, while the remaining binarizers form a clearly lower-performing group.

The HBF1 results in Tab.~\ref{tab:results_hbf1} show a different aspect of the forensic setting. In the controlled Blood-vs-Non-Blood task, nearly all methods reach saturated performance, indicating that this scene provides a comparatively clean source of blood spectra. Differences are more visible in the multiclass task, where SHB remains among the strongest representations: TM-SHB matches the best Balanced Accuracy, while RF-SHB stays within the top group. Thus, unlike the more challenging HBE1 scene, HBF1 mainly confirms that SHB does not lose performance in an easier laboratory scenario.

The INPI results in Tab.~\ref{tab:results_inpi} show that the same pattern also holds in a classic remote-sensing vegetation benchmark. SHB-based methods lead all evaluated tasks: RF-SHB is strongest in the multiclass setting, while TM-SHB ranks first in all three binary vegetation comparisons. The gains are especially visible in the crop-type pairs, where SHB substantially improves over RF-Raw and over the conventional binarisation baselines, suggesting that spectral contrast features are well suited to separating vegetation classes with similar overall reflectance profiles.

The MOSA results in Tab.~\ref{tab:results_mosa} confirm the trend on a particularly demanding rainforest dataset, where species-level canopy discrimination is substantially harder than the more controlled forensic blood tasks. RF-SHB clearly leads the multiclass setting and the BAEL-vs-DIPA task, while TM-SHB achieves the best result for LEAM-vs-CEPE. In all cases, SHB improves over RF-Raw and over the conventional binarisation baselines, suggesting that spectral contrast features are especially useful when subtle differences between vegetation spectra must be extracted from realistic, heterogeneous canopy data.

The PAVI results in Tab.~\ref{tab:results_pavi} extend the same conclusion to an urban remote-sensing scene, where spectral signatures correspond to materials such as asphalt, bitumen, meadows, trees, and built surfaces. SHB remains among the strongest representations across all tasks: RF-SHB leads the multiclass setting, while TM-SHB obtains the best Balanced Accuracy in both binary comparisons. The gains are smaller than in INPI or MOSA for some tasks, especially where RF-Raw is already competitive, but SHB consistently ranks at the top, indicating that the proposed spectral contrasts are useful not only for vegetation spectra but also for urban material discrimination.

\subsubsection{Statistical Comparison}

\begin{table*}[htbp]
\centering
\caption{Comparison between TM-SHB pipeline and reference models in binary and multiclass tasks, using Bayesian methodology from~\cite{benavoli2017time}. Probabilities denote whether the control method is strictly better ($P(C > B)$), worse ($P(B > C)$), practically equivalent ($P(C \approx B)$). The value of ROPE is set to 1\%}
\begin{tabular}{lllrrr}
\toprule
Task Type & Control ($C$) & Baseline ($B$) & $P(C > B)$ & $P(C \approx B)$ & $P(B > C)$ \\
\midrule
\multirow{6}{*}{Binary} & \multirow{6}{*}{TM-SHB} & RF-SHB & 8.34\% & 91.08\% & 0.57\% \\
 &  & TM-Std & 99.55\% & 0.00\% & 0.45\% \\
 &  & RF-Std & 99.84\% & 0.00\% & 0.16\% \\
 &  & RF-Raw & 99.46\% & 0.00\% & 0.54\% \\
 &  & TM-MultiOtsu-Thermometer & 99.72\% & 0.00\% & 0.28\% \\
 &  & TM-KMeans-Thermometer & 99.77\% & 0.00\% & 0.23\% \\
\midrule
\multirow{6}{*}{Multiclass} & \multirow{6}{*}{TM-SHB} & RF-SHB & 9.91\% & 0.13\% & 89.96\% \\
 &  & TM-Std & 98.41\% & 0.00\% & 1.59\% \\
 &  & RF-Std & 98.72\% & 0.00\% & 1.28\% \\
 &  & RF-Raw & 28.57\% & 51.73\% & 19.70\% \\
 &  & TM-MultiOtsu-Thermometer & 97.62\% & 0.00\% & 2.38\% \\
 &  & TM-KMeans-Thermometer & 97.62\% & 0.00\% & 2.38\% \\
\bottomrule
\end{tabular}
\label{tab:probabilities}
\end{table*}

\begin{figure}
    \centering
    \begin{subfigure}{0.45\linewidth}
        \centering
        \includegraphics[width=\linewidth]{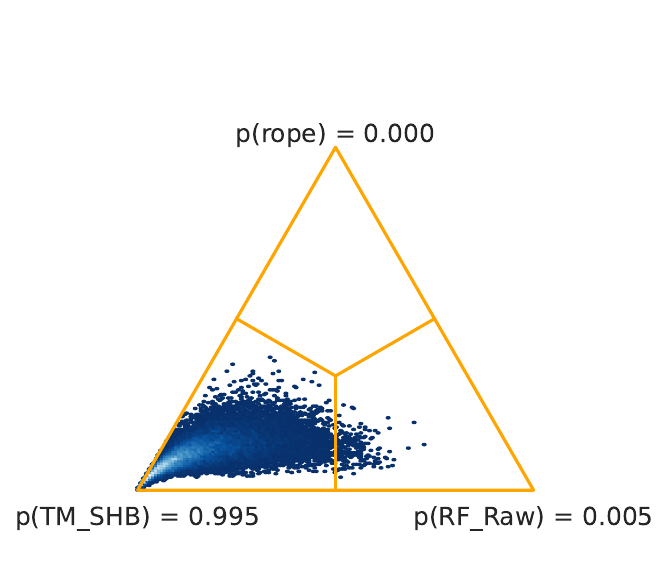}
        \caption{Binary, TM-SHB vs RF-RAW}
    \end{subfigure}
    \hfill
    \begin{subfigure}{0.45\linewidth}
        \centering
        \includegraphics[width=\linewidth]{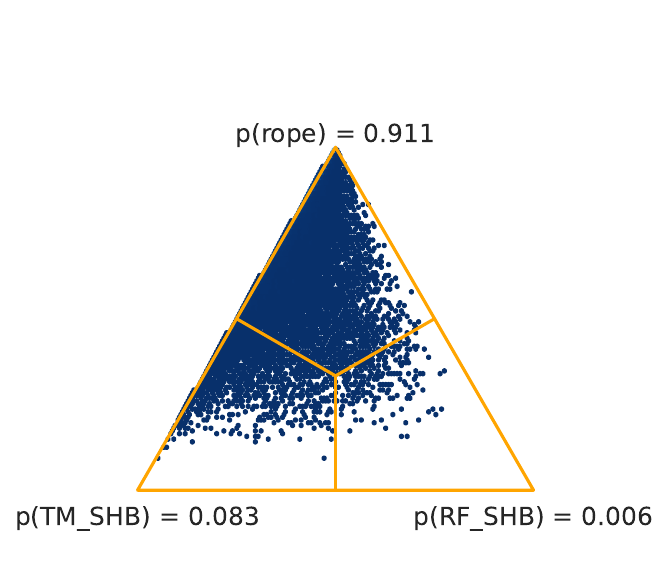}
        \caption{Binary, TM-SHB vs RF-SHB}
    \end{subfigure}
    \begin{subfigure}{0.45\linewidth}
        \centering
        \includegraphics[width=\linewidth]{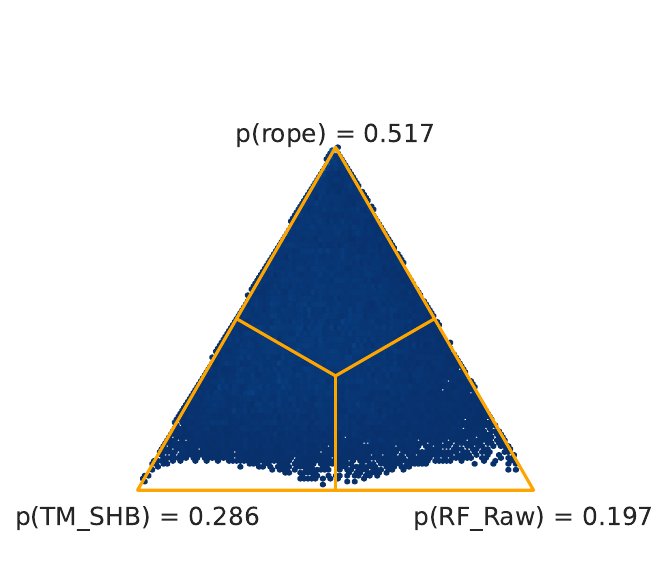}
        \caption{Multiclass, TM-SHB vs RF-RAW}
    \end{subfigure}
    \hfill
    \begin{subfigure}{0.45\linewidth}
        \centering
        \includegraphics[width=\linewidth]{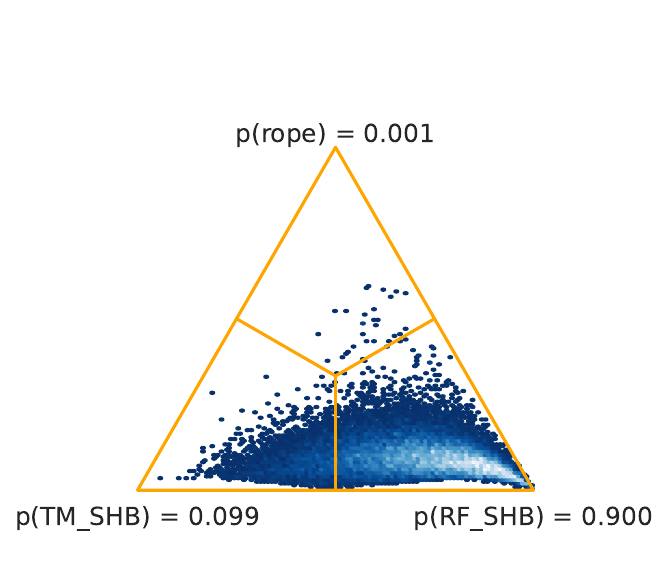}
        \caption{Multiclass, TM-SHB vs RF-SHB}
    \end{subfigure}
    \caption{Visualization of probability that TM-SHB is better, worse or practically equivalent, indicated by p(rope), than selected reference, based on Bayesian methodology from~\cite{benavoli2017time}. Note that compared classifiers use either the proposed binarisation scheme or continuous data, without binarisation.}
    \label{fig:bayes}
\end{figure}

Table~\ref{tab:probabilities} reports the Bayesian statistical comparison between the proposed TM-SHB pipeline and selected reference methods. The results show that TM-SHB has a very high posterior probability of outperforming the traditional TM-compatible binarisation schemes in both binary and multiclass settings. This is observed for the Standard Binarizer, Multi-Otsu, and K-Means variants, where $P(C>B)$ is above $97\%$ in all reported comparisons. This confirms that the proposed spectral contrast representation provides a substantially stronger Boolean input space for Tsetlin Machines than other discretisation approaches.

The comparison with RF-based references gives a more nuanced picture. In binary tasks, TM-SHB is practically equivalent to RF-SHB with high probability ($P(C \approx B)=91.08\%$), and strongly outperforms RF-Raw ($P(C>B)=99.46\%$). This is notable because RF-Raw operates directly on the continuous spectra, whereas TM-SHB uses a Boolean representation. In multiclass tasks, RF-SHB is favoured over TM-SHB ($P(B>C)=89.96\%$), while TM-SHB and RF-Raw are most probably practically equivalent ($P(C \approx B)=51.73\%$). These findings are consistent with the probability visualizations in Fig.~\ref{fig:bayes} and with the averaged scores in Table~\ref{tab:performance}: SHB is consistently beneficial, but RF-SHB remains stronger than TM-SHB on the most difficult multiclass configurations.
\subsubsection{Ranking Comparison}

\begin{table*}
\centering
\caption{Ranking table for tested binariser-classifier combination, in binary and multiclass experiments. Models with the highest number of points performed better across datasets.}
\begin{tabular}{lcccccc}
\toprule
Method & \multicolumn{3}{c}{Macro-F1} & \multicolumn{3}{c}{Bal-Acc} \\
\cmidrule(lr){2-4} \cmidrule(lr){5-7}
 & Bin. & Multi. & Total & Bin. & Multi. & Total \\
\midrule
RF-SHB & 26 & 15 & 41 & 24 & 15 & 39 \\
TM-SHB & 26 & 10 & 36 & 26 & 12 & 38 \\
RF-Raw & 17 & 12 & 29 & 15 & 11 & 26 \\
RF-Std & 12 & 4 & 16 & 9 & 4 & 13 \\
RF-MultiOtsu-Thermometer & 10 & 5 & 15 & 7 & 5 & 12 \\
RF-Quantile-Thermometer & 8 & 4 & 12 & 6 & 4 & 10 \\
RF-MultiOtsu-OneHot & 6 & 5 & 11 & 6 & 4 & 10 \\
TM-MultiOtsu-OneHot & 7 & 3 & 10 & 5 & 3 & 8 \\
TM-MultiOtsu-Thermometer & 7 & 3 & 10 & 4 & 2 & 6 \\
RF-KMeans-Thermometer & 5 & 4 & 9 & 6 & 3 & 9 \\
RF-KMeans-OneHot & 4 & 5 & 9 & 5 & 3 & 8 \\
RF-Quantile-OneHot & 5 & 4 & 9 & 6 & 4 & 10 \\
TM-Std & 7 & 1 & 8 & 6 & 0 & 6 \\
TM-Quantile-Thermometer & 6 & 2 & 8 & 8 & 1 & 9 \\
TM-KMeans-OneHot & 4 & 3 & 7 & 3 & 3 & 6 \\
TM-KMeans-Thermometer & 3 & 3 & 6 & 3 & 3 & 6 \\
TM-Quantile-OneHot & 4 & 2 & 6 & 4 & 1 & 5 \\
\bottomrule
\end{tabular}
\label{tab:ranks}
\end{table*}

Table~\ref{tab:ranks} summarises the results from a complementary rank-based perspective. The SHB-based methods again dominate this comparison: RF-SHB obtains the highest total score for both Macro-F1 and Balanced Accuracy, while TM-SHB ranks second overall and achieves the strongest binary-task score in Balanced Accuracy. This indicates that SHB is not only competitive in average performance but also consistently ranks among the best-performing configurations on individual tasks.

The ranking also highlights the strength of RF-Raw as a non-binarised reference, which ranks third overall. However, both SHB-based configurations outperform it in the aggregate scores, suggesting that the proposed spectral contrast representation provides a more effective input space than the raw spectra for the evaluated settings. The remaining reference binarisation methods obtain substantially lower totals, with RF-based variants generally ranking above their TM counterparts. Overall, the rank-based analysis supports the conclusions from the conventional metrics and Bayesian comparison: SHB provides the most robust representation across datasets, metrics, and task types.

\section{Discussion}
Across the evaluated datasets, tasks, classifiers, and metrics, the results consistently show that the proposed Spectral Haar Binarizer (SHB) provides a substantially stronger representation than the reference binarisation methods. SHB-based configurations achieve the best average performance in both binary and multiclass settings, dominate the rank-based summary, and are strongly favoured by the Bayesian comparison against standard per-band discretisation schemes. The strongest binary results are obtained by TM-SHB, indicating that the proposed Boolean spectral contrast features are well aligned with Tsetlin Machine clauses. In multiclass problems, RF-SHB is generally the strongest configuration, while TM-SHB remains the best TM-based method and is practically competitive with the RF classifier trained on raw spectra. Importantly, these trends hold across markedly different hyperspectral domains, including forensic blood detection, agricultural and urban remote sensing, and rainforest tree-species classification, suggesting that SHB captures reusable spectral structure rather than dataset-specific artifacts.

The following discussion interprets these findings from selected complementary perspectives.

\subsection{Effect of Binarisation Resolution}\label{sec:ablation_bits}
\begin{figure}
    \centering
    \begin{subfigure}{0.8\linewidth}
        \centering
        \includegraphics[width=\linewidth]{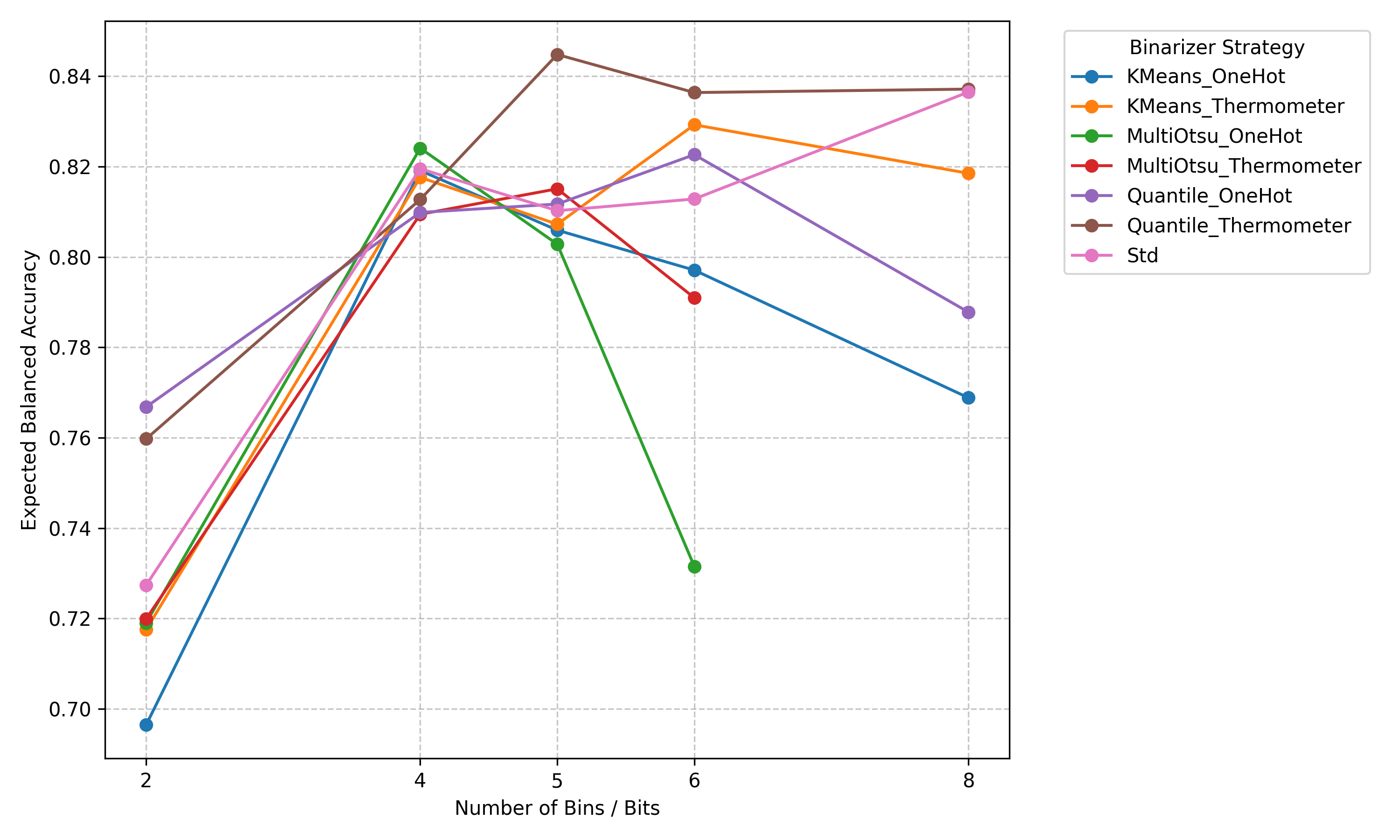}
        \caption{Binary}
    \end{subfigure}
    \hfill
    \begin{subfigure}{0.8\linewidth}
        \centering
        \includegraphics[width=\linewidth]{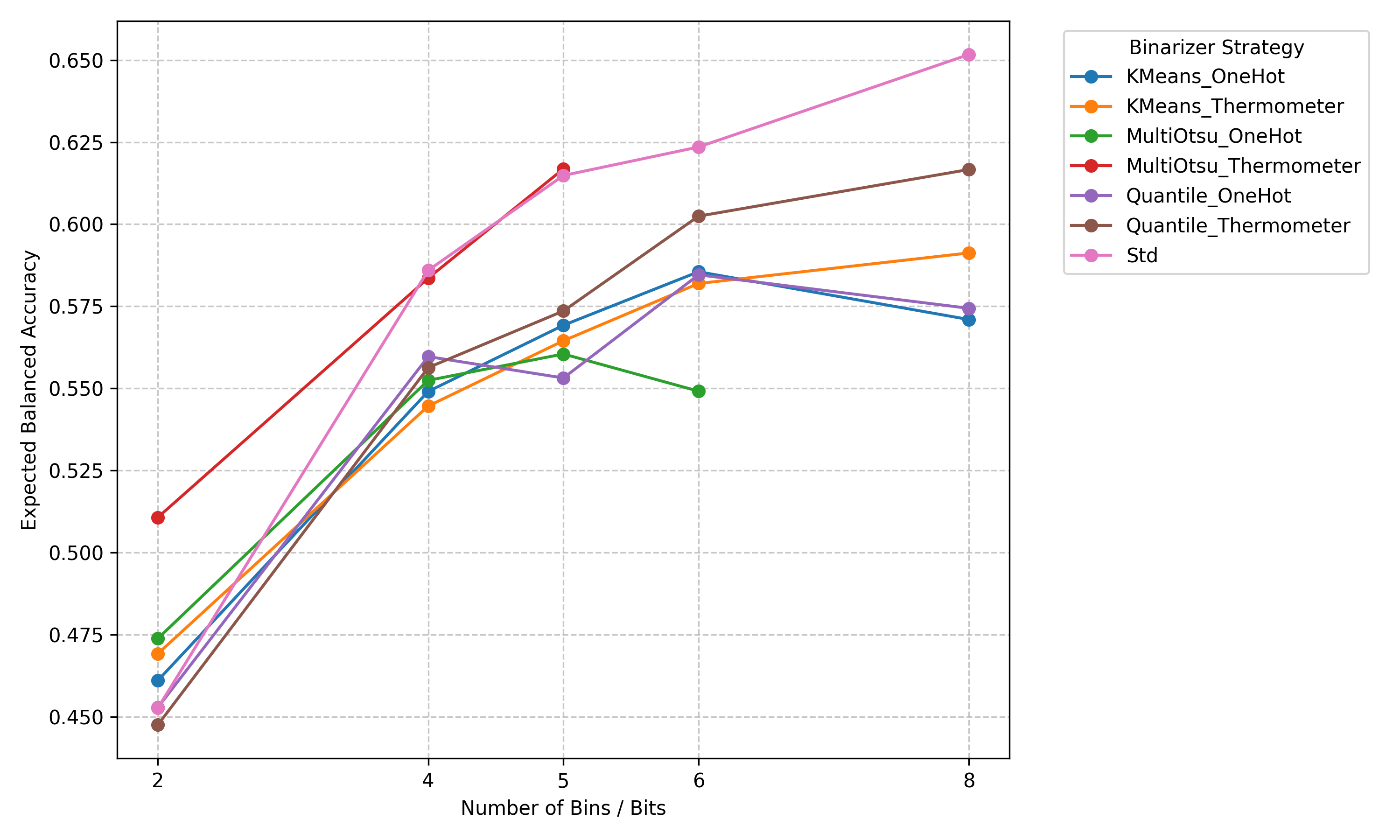}
        \caption{Multiclass}
    \end{subfigure}
    \begin{subfigure}{0.8\linewidth}
        \centering
        \includegraphics[width=\linewidth]{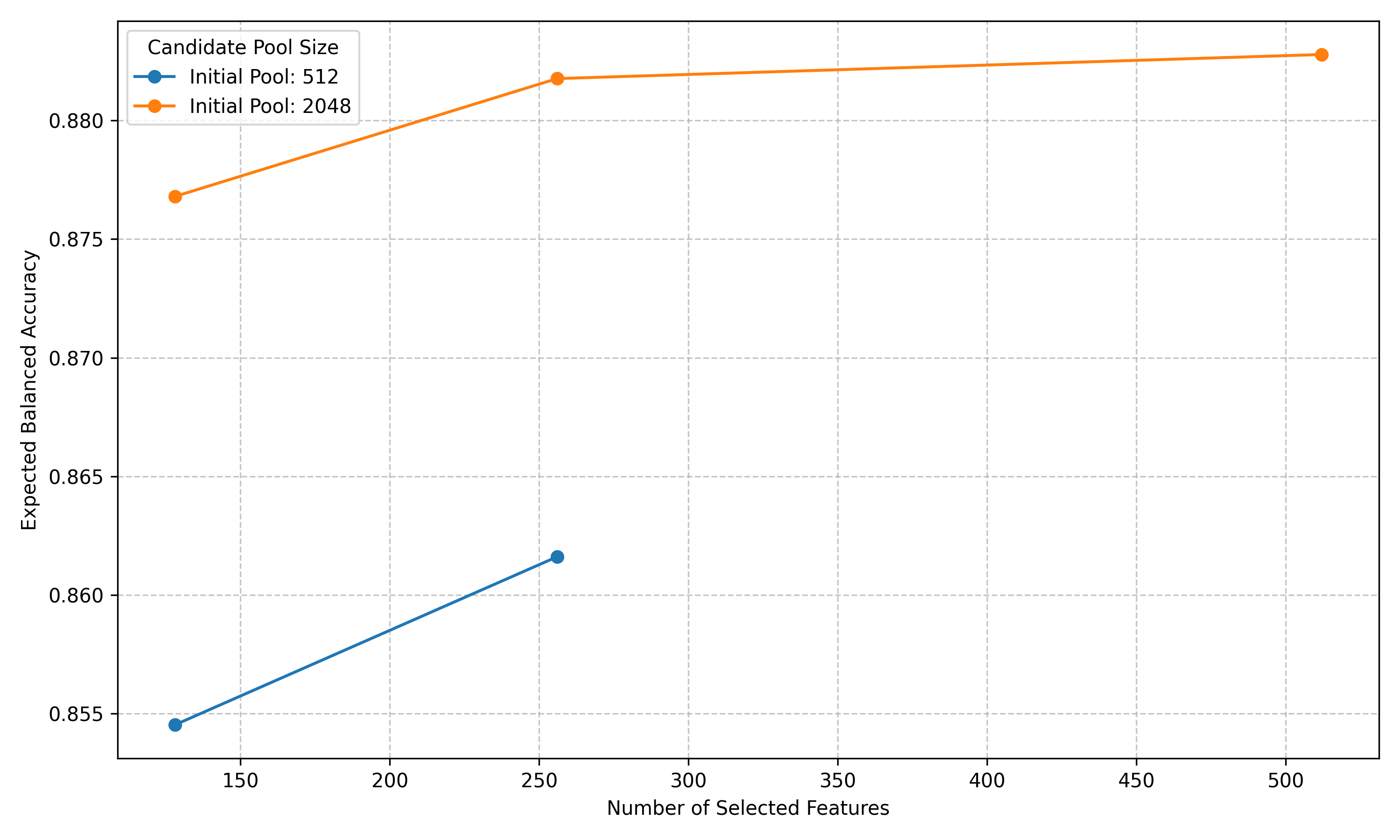}
        \caption{SHB features}
    \end{subfigure}
    \hfill
    \caption{Ablation study: visualization of the impact of the number of bits parameter on binary (a) and multiclass (b) balanced accuracy. Panel (c) presents the impact of the number of features (filters) on the SHB method.}
    \label{fig:ablation}
\end{figure}

To determine the configuration of the reference binarizers, we conducted a targeted ablation study evaluating the impact of binarisation resolution on classification performance. For the traditional discretisation methods, namely Standard, K-Means, Quantile, and Multi-Otsu, we varied the bit depth, i.e., the number of bins per continuous feature, from 2 to 8. For the proposed Spectral Haar Binarizer (SHB), we evaluated performance across different sizes of the selected feature set.

The inherent trade-off in discretisation is that increasing resolution improves the precision with which continuous values are represented but also increases the dimensionality of the resulting Boolean space. Since hyperspectral data is already high-dimensional at the source, representing each spectral channel with many binary features quickly becomes computationally expensive and substantially increases the memory footprint.

As illustrated in Fig.~\ref{fig:ablation}, our results indicate that, for binary classification tasks, Balanced Accuracy generally peaks around 4--5 bits before plateauing or slightly degrading. In multiclass scenarios, performance shows a modest continued increase at higher bit depths, but the gains become progressively smaller. Balancing computational efficiency against classification precision, we therefore selected 5 bits per feature as a sensible middle ground.

Similarly, for SHB, classification performance improves as the number of selected features increases, as shown in Fig.~\ref{fig:ablation}(c). However, performance begins to stabilise after approximately 256 selected features. Consequently, we fixed the SHB configuration to extract 256 features for all subsequent evaluations, using 1024 candidates as the initial feature pool. This provides a compact yet discriminative binary representation while keeping the feature-selection process computationally manageable.

\subsection{Compactness of the Binary Representation}\label{sec:featuresize}

For the Standard Binarizer, the length of the resulting binary vector is determined by the number of spectral features and the number of bits assigned to each feature. Specifically, for $F$ spectral channels and $n_{\mathrm{bit}}$ bits per channel, the output dimensionality is
\[
    D = F \times n_{\mathrm{bit}}.
\]
Thus, for $F=200$ spectral channels and $n_{\mathrm{bit}}=5$, the binarised representation expands to $1000$ Boolean features.

The same dimensionality scaling applies to bin-based encodings such as K-Means, Quantile, and Multi-Otsu when one-hot encoding is used. If each of the $F$ spectral channels is discretised into $k$ mutually exclusive bins, the one-hot representation contains $k$ Boolean variables per channel, yielding
\[
    D = F \times k.
\]
For $k=5$ bins and $F=200$ channels, this again gives a binary vector of length $1000$.

In contrast, the Spectral Haar Binarizer decouples the dimensionality of the binary representation from the raw spectral dimensionality $F$. Its output length is determined by the number of selected spectral contrast features, denoted by $S$. A configuration with $S=256$ therefore produces a binary vector of exactly $256$ features, regardless of the initial number of spectral bands. In this setting, SHB yields approximately a fourfold reduction in representation size relative to the $1000$-dimensional outputs produced by the reference binarizers with $5$ bits or bins per channel.

This reduction is important because the size of the Boolean input directly affects memory usage and the cost of subsequent TM inference. The effect can be even stronger when early stopping is enabled during feature selection, since fewer than $S$ features may be retained. However, as shown in the ablation study, overly aggressive representation reduction can lead to a loss of classification accuracy.

\subsection{Results in the Context of Relevant Literature}

The obtained results are consistent with a long-standing observation in hyperspectral image analysis: the effectiveness of a classifier depends strongly on the representation used to expose spectral information to the learning algorithm. Classical studies have shown that high-dimensional hyperspectral spectra can be classified effectively by strong general-purpose classifiers such as SVMs and Random Forests~\cite{Melgani2004svm,Pal2005random,Joelsson2005random}, while more recent reviews emphasize that feature extraction remains central to HSI classification despite the growth of deep spectral--spatial models~\cite{Li2019deep,Rasti2020feature,Grewal2023survey}. In this context, the strong performance of RF-Raw in our experiments is expected: Random Forests are robust classifiers for high-dimensional tabular data and provide a demanding non-binarised baseline. The fact that SHB improves over RF-Raw in several settings and remains competitive in others suggests that the proposed spectral contrast representation captures useful structure rather than merely compressing the input.

Our results also connect directly to the Tsetlin Machine literature on continuous inputs. Standard TMs operate on Boolean variables~\cite{granmo2018tsetlin}, and previous work has therefore introduced threshold-based schemes that convert continuous features into Boolean predicates~\cite{abeyrathna2019continuous}. The adaptive sparse representation of continuous TM inputs further shows that the design of this Boolean interface is not merely a secondary implementation detail but a central modelling choice that affects both expressiveness and dimensionality~\cite{abeyrathna2021adaptive}. The performance gap observed between SHB and the reference binarizers empirically supports this point for hyperspectral data: simple per-band discretisation yields a valid Boolean input space, but it does not capture spectral relationships as effectively as contrast features that span local intervals of the spectrum.

The proposed method is also related to earlier attempts to adapt TMs to structured non-binary sensor data. The Convolutional Tsetlin Machine demonstrated that TM clauses can be extended to image-like inputs through local binary patterns~\cite{granmo2019convolutional}, while MixCTME used Otsu-binarized spectrogram representations before classification with convolutional TM experts~\cite{jeeru2025mixctme}. Our work follows the same general direction, but shifts the emphasis from spatial or time--frequency images to one-dimensional hyperspectral signatures. In this sense, SHB can be viewed as a spectral analogue of local contrast features: instead of comparing rectangular image regions as in Viola--Jones~\cite{Viola2004robust}, it compares selected wavelength intervals and converts the resulting contrasts into Boolean predicates.

Finally, the results are relevant to the broader motivation behind compact, hardware-friendly classifiers. Recent TM hardware studies show that Boolean clause evaluation and bit-level operations are attractive for efficient inference and on-device learning~\cite{Tunheim2025fpga,Mao2025dynamic}. Our findings complement that line of work by addressing the representation problem that arises before such hardware advantages can be exploited for continuous sensor data. If hyperspectral spectra can be converted into compact, discriminative Boolean features without substantial loss of accuracy, then TMs become more realistic candidates for resource-constrained sensing scenarios, including onboard or near-sensor classification. Thus, the contribution of SHB is not only improved accuracy over baseline binarisation schemes but also a practical bridge between continuous hyperspectral measurements and the Boolean logic required by efficient Tsetlin Machine pipelines.

\subsection{Why Spectral Contrast Features Help}

The main advantage of SHB appears to come from changing the unit of binarisation. Reference binarizers operate independently on individual spectral bands: each Boolean variable indicates whether a single wavelength value exceeds a threshold, falls within a bin, or lies within a data-driven quantisation interval. Such predicates are simple and natural for Tsetlin Machines, but they treat each band largely in isolation. This is a limiting assumption for hyperspectral data, where class-discriminative information is often not contained in the absolute value of a single band, but in the relative shape of the spectrum: local slopes, absorption-like depressions, shoulders, and contrasts between neighbouring or partially overlapping wavelength regions.

SHB exposes precisely this type of information to the classifier. By comparing aggregated responses over pairs of spectral intervals, it converts local spectral shape into Boolean predicates before TM clause learning begins. A selected feature may therefore express conditions such as whether one spectral region is consistently stronger than another, whether a narrow interval deviates from its surrounding context, or whether a broader spectral trend differs between classes. These are still simple binary tests, but they are defined in terms of relationships between bands rather than isolated band values. This makes the Boolean input space more aligned with the structure of hyperspectral measurements.

This also helps explain why SHB improves both TM and RF configurations. For Tsetlin Machines, the benefit is direct: clauses can combine more informative spectral predicates, reducing the need to reconstruct band relationships from many independent threshold variables. For Random Forests, SHB serves as a compact feature-extraction stage, replacing raw or independently discretised spectra with contrast features that encode relevant spectral comparisons. The fact that SHB is beneficial for both classifiers suggests that the gain is due not only to better compatibility with TMs but also to the representational value of the spectral contrasts themselves.

At the same time, SHB does not abandon the advantages of Boolean representation. The selected contrast predicates remain binary, compact, and interpretable at the level of spectral regions. This is important because it preserves the computational and logical structure required by Tsetlin Machines while avoiding the information loss associated with overly local per-band discretisation. In this sense, SHB can be viewed as a compromise between hand-crafted spectral feature extraction and TM-compatible binarisation: it introduces domain-aware spectral comparisons while keeping the final representation simple enough for efficient Boolean clause learning.

The results also show that the effect of SHB depends on the downstream classifier. For Tsetlin Machines, SHB directly addresses the central bottleneck of continuous-input binarisation: instead of forcing clauses to build spectral relationships from many independent per-band thresholds, it provides Boolean predicates that already encode contrasts between spectral regions. This explains the large improvement of TM-SHB over the other TM-compatible binarisation schemes. For Random Forests, SHB acts more like a compact feature-extraction stage. RF-SHB is particularly strong in multiclass settings, indicating that the selected spectral contrasts are useful beyond the TM architecture itself. Rather than weakening the role of TMs, this result supports the broader interpretation that SHB captures discriminative hyperspectral structure in a classifier-agnostic way, while remaining especially valuable for TMs by providing this structure in Boolean form.

\subsection{Robustness Across Hyperspectral Domains}

An important aspect of the experimental results is that the observed advantage of SHB is not limited to a single type of hyperspectral scene. The evaluation includes close-range forensic imaging, agricultural and urban remote sensing, and tropical forest species discrimination. These domains differ substantially in acquisition geometry, spatial resolution, spectral range, class semantics, and the physical interpretation of the measured spectra. Despite this heterogeneity, SHB remains among the strongest representations across all datasets. This suggests that the method is not exploiting a dataset-specific artifacts, such as a particular wavelength range, sensor response, or class distribution. Instead, the results point to a more general property of hyperspectral data: class information is often expressed through relative spectral structure rather than through isolated band values alone. By constructing predicates from contrasts between spectral intervals, SHB captures information relevant across different sensing scenarios.

The domain-specific results also show that the magnitude and character of the gain depend on the task. In simpler binary settings, especially when the classes are already well-separated, several methods can achieve high performance, and the advantage of SHB is correspondingly smaller. In more difficult settings, such as multiclass classification or fine-grained vegetation discrimination, the benefit of a more structured representation becomes more visible. This pattern is consistent with the interpretation that SHB is most useful when discrimination depends on subtle differences in spectral shape, rather than on large absolute differences in reflectance or absorbance.

At the same time, the results should not be read as evidence of universal domain invariance. SHB was evaluated on a deliberately diverse, but still finite, set of hyperspectral datasets. Further validation would be needed for sensors with different spectral coverage, strongly varying atmospheric or illumination conditions, and application domains with different noise characteristics. Nevertheless, the present benchmark provides evidence that contrast-based Boolean spectral features can transfer across substantially different hyperspectral problems, making SHB a promising general-purpose binarisation strategy rather than a narrowly tuned preprocessing step.

\subsection{Implementation Considerations}

Several implementation choices were made to keep the proposed SHB pipeline simple, reproducible, and computationally manageable. First, candidate spectral-contrast features are randomly generated subject to constraints on interval layout and maximum window size. This provides broad coverage of possible interval pairs without exhaustive enumeration. However, purely random sampling may underrepresent structures that are particularly meaningful in hyperspectral data, such as narrow absorption-like bands, local slope changes, or contrasts between short and broad spectral regions. A natural extension would be to retain stochastic generation while biasing the candidate pool toward planned spectral layouts, including adjacent intervals at multiple scales and narrow-vs-broad comparisons. This could increase the likelihood of generating informative filters and reduce the required candidate pool size.

Second, feature selection in multiclass problems is handled through a one-vs-rest scheme with boosting rounds distributed approximately evenly across classes. This ensures that every class contributes to the final representation, but it also assumes that all classes require a similar feature budget. In practice, some classes may be spectrally easier to separate than others. Future versions could enforce only a small minimum quota per class and allocate the remaining feature budget adaptively, for example, according to validation error, class frequency, or the margins observed during early boosting rounds.

Finally, the current implementation includes several pragmatic safeguards to improve efficiency and avoid redundant predicates. Duplicate interval candidates are removed before feature selection, exact threshold search is performed using pre-sorted feature responses, and previously selected stumps are excluded from later boosting rounds. In the multiclass setting, duplicate predicates selected by different one-vs-rest subproblems are merged by retaining the instance with the highest stump importance. Optional early stopping can also terminate feature selection before the requested number of features is reached, based on weak stump importance, stagnating weighted error, or collapse of the sample-weight distribution. These choices make the method more stable and efficient, but they are intentionally conservative: no correlation-based pruning or geometry-aware deduplication is currently performed. Such refinements remain promising avenues for further reducing representation size without sacrificing classification accuracy.

\subsection{Limitations and Future Work}

Although the benchmark used in this study is deliberately heterogeneous, it still covers only a subset of possible hyperspectral operating conditions. Future work should evaluate SHB on domains with different physical signal formation and noise characteristics, including low-SNR airborne or satellite scenes, strongly illumination-dependent field measurements, biomedical hyperspectral data, mineralogical or chemical mapping, and astrophysical or planetary spectral cubes. Such datasets may contain sharper emission or absorption lines, stronger calibration artifacts, atmospheric effects, missing bands, or much higher noise levels than the datasets considered here. Testing SHB under these conditions would clarify whether interval-based spectral contrasts remain robust when the spectral structure is less smooth, less calibrated, or governed by different physical mechanisms.

A second limitation is that the present study operates on individual hyperspectral vectors. This design isolates the effect of spectral binarisation and makes the comparison between binarizers clean, but it does not exploit spatial neighbourhoods. In hyperspectral image classification, spatial--spectral context is known to improve performance, especially in remote-sensing scenes where neighbouring pixels often belong to the same material or land-cover class. A natural extension is therefore to combine SHB with spatial neighbourhoods, morphological profiles, patch-based representations, or convolutional Tsetlin Machine variants. This should, however, be evaluated carefully, since spatial processing can also introduce optimistic results if train--test splits are not designed to control spatial overlap.

Several implementation-level refinements also remain open. Candidate spectral contrasts are currently generated randomly under structural constraints, which provides broad coverage but may undersample physically meaningful patterns such as narrow absorption-like bands, adjacent multi-scale contrasts, or narrow-vs-broad interval comparisons. Future versions could bias the candidate pool toward such layouts while retaining stochastic diversity. Similarly, the current one-vs-rest multiclass feature-selection procedure allocates the feature budget approximately evenly across classes, although some classes are clearly harder to separate than others. More adaptive allocation strategies, based on validation error, class frequency, or early boosting margins, could make the selected representation more efficient. Finally, the representation size is fixed globally in the main experiments. Although the ablation study supports the choice of $S=256$, future work should investigate adaptive feature budgets selected per dataset, per task, or even per class.

The present evaluation also focuses on predictive performance rather than full deployment. Since one motivation for combining SHB with Tsetlin Machines is efficient Boolean inference, a necessary next step is an actual field or near-field deployment study. Such a study should include the complete pipeline: sensor acquisition, preprocessing, SHB feature extraction, TM inference, and post-processing under realistic memory, latency, and energy constraints. Hardware-aware optimisation is also important. SHB reduces the dimensionality of the Boolean representation, but the costs of feature generation, thresholding, memory access, and clause evaluation should be measured on edge devices, embedded platforms, or FPGA implementations rather than inferred solely from the representation size.

Finally, the interpretability of SHB features has not yet been fully exploited. Each selected feature corresponds to a contrast between two spectral intervals, which makes it more interpretable than an arbitrary latent feature. However, the present study does not systematically confront these selected intervals with domain knowledge. Future work should analyse whether frequently selected SHB features correspond to known absorption bands, material-specific spectral regions, vegetation characteristics, blood-related spectral features, or other physically meaningful signatures. Such analysis should ideally involve domain specialists, for example, forensic experts, remote-sensing analysts, ecologists, mineralogists, or astronomers, depending on the application. This would allow SHB to be evaluated not only as an accurate binarisation method, but also as a tool for discovering and communicating meaningful spectral evidence.

\section{Conclusions}
This paper introduced the Spectral Haar Binarizer (SHB), a contrast-based binarisation method that converts continuous hyperspectral spectra into compact Boolean representations for Tsetlin Machine classification. Instead of discretising each spectral band independently, SHB constructs binary predicates from contrasts between selected spectral intervals, thereby exposing local and regional spectral structure to Boolean clause learning.

Across five heterogeneous hyperspectral datasets, SHB consistently outperformed standard per-band binarisation strategies and produced the strongest or near-strongest results across binary and multiclass tasks. In particular, TM-SHB achieved the best average performance in binary classification, while RF-SHB was strongest in multiclass settings. Bayesian and rank-based analyses further confirmed that the advantage of SHB is robust across datasets and not driven by isolated experimental cases.

The results show that binarisation is not merely a technical preprocessing step for Tsetlin Machines, but a central representation-learning problem. By providing compact, interpretable, and discriminative Boolean features, SHB offers a practical bridge between information-rich hyperspectral measurements and efficient logic-based classifiers. Future work should extend the method to spatial--spectral neighbourhoods, broader sensor domains, hardware-oriented deployment, and expert-driven interpretation of selected spectral contrasts.

\section*{Acknowledgments}
The authors acknowledge the use of AI-based tools, such as ChatGPT and Grammarly, for assistance in editing, grammar enhancement, and spelling checks during the preparation of this manuscript.

\appendix
\section{Supplementary results}
Comprehensive experimental results for evaluated datasets are presented in Table~\ref{tab:results_hbe1} (HBE1), Table~\ref{tab:results_hbf1} (HBF1), Table~\ref{tab:results_inpi} (INPI), Table~\ref{tab:results_mosa} (MOSA), Table~\ref{tab:results_pavi} (PAVI).

\begin{table*}[htbp]
\centering
\caption{Comprehensive experimental results for the HBE1 dataset (Mean $\pm$ Standard Deviation across 5 CV folds). Ordered by BAcc.}
\label{tab:results_hbe1}
\begin{tabular}{llrrr}
\toprule
Task / Pair & Method & Test Acc & Test BAcc & Test F1 \\
\midrule
\multirow{17}{*}{Multiclass} & RF-SHB & $98.73 \pm 0.16$ & $97.66 \pm 0.24$ & $97.83 \pm 0.22$ \\
 & TM-SHB & $98.31 \pm 0.33$ & $97.04 \pm 0.48$ & $94.86 \pm 5.28$ \\
 & RF-Raw & $97.25 \pm 0.15$ & $96.13 \pm 0.18$ & $96.23 \pm 0.18$ \\
 & RF-Quantile-OneHot & $92.44 \pm 0.15$ & $90.26 \pm 0.28$ & $90.36 \pm 0.21$ \\
 & RF-Quantile-Thermometer & $92.41 \pm 0.10$ & $90.25 \pm 0.18$ & $90.37 \pm 0.15$ \\
 & RF-Std & $92.36 \pm 0.16$ & $90.13 \pm 0.24$ & $90.25 \pm 0.21$ \\
 & RF-MultiOtsu-OneHot & $92.01 \pm 0.04$ & $89.64 \pm 0.15$ & $89.93 \pm 0.09$ \\
 & RF-MultiOtsu-Thermometer & $91.97 \pm 0.18$ & $89.58 \pm 0.35$ & $89.92 \pm 0.30$ \\
 & RF-KMeans-OneHot & $91.20 \pm 0.33$ & $88.86 \pm 0.31$ & $89.01 \pm 0.35$ \\
 & RF-KMeans-Thermometer & $91.10 \pm 0.17$ & $88.76 \pm 0.15$ & $88.93 \pm 0.19$ \\
 & TM-Quantile-OneHot & $90.04 \pm 0.33$ & $87.30 \pm 0.55$ & $78.83 \pm 5.08$ \\
 & TM-MultiOtsu-OneHot & $89.66 \pm 0.30$ & $86.68 \pm 0.50$ & $87.29 \pm 0.42$ \\
 & TM-Quantile-Thermometer & $89.35 \pm 0.61$ & $86.12 \pm 0.65$ & $84.66 \pm 4.82$ \\
 & TM-MultiOtsu-Thermometer & $88.73 \pm 0.56$ & $85.47 \pm 1.12$ & $84.06 \pm 4.73$ \\
 & TM-KMeans-OneHot & $88.40 \pm 0.25$ & $85.34 \pm 0.26$ & $81.58 \pm 5.91$ \\
 & TM-KMeans-Thermometer & $87.72 \pm 0.35$ & $84.56 \pm 0.75$ & $85.23 \pm 0.41$ \\
 & TM-Std & $88.37 \pm 0.18$ & $84.56 \pm 0.35$ & $85.49 \pm 0.24$ \\
\midrule
\multirow{17}{*}{Blood-vs-Non-Blood} & TM-SHB & $99.17 \pm 0.13$ & $97.11 \pm 0.40$ & $97.24 \pm 0.42$ \\
 & RF-SHB & $99.09 \pm 0.10$ & $96.33 \pm 0.45$ & $96.95 \pm 0.35$ \\
 & RF-Raw & $98.82 \pm 0.11$ & $94.29 \pm 0.38$ & $95.94 \pm 0.36$ \\
 & RF-Quantile-OneHot & $96.71 \pm 0.17$ & $87.23 \pm 1.33$ & $88.68 \pm 0.73$ \\
 & RF-Std & $96.68 \pm 0.13$ & $87.11 \pm 1.44$ & $88.59 \pm 0.62$ \\
 & RF-Quantile-Thermometer & $96.65 \pm 0.14$ & $87.04 \pm 1.06$ & $88.50 \pm 0.49$ \\
 & RF-MultiOtsu-OneHot & $96.25 \pm 0.06$ & $84.90 \pm 0.94$ & $86.89 \pm 0.29$ \\
 & RF-MultiOtsu-Thermometer & $96.13 \pm 0.10$ & $84.29 \pm 0.89$ & $86.42 \pm 0.33$ \\
 & RF-KMeans-OneHot & $96.04 \pm 0.09$ & $84.08 \pm 1.50$ & $86.11 \pm 0.63$ \\
 & RF-KMeans-Thermometer & $96.00 \pm 0.13$ & $83.57 \pm 1.42$ & $85.86 \pm 0.74$ \\
 & TM-Quantile-Thermometer & $95.07 \pm 1.08$ & $82.49 \pm 1.96$ & $83.41 \pm 2.01$ \\
 & TM-MultiOtsu-OneHot & $95.24 \pm 0.13$ & $78.21 \pm 2.65$ & $82.00 \pm 1.38$ \\
 & TM-KMeans-OneHot & $94.96 \pm 0.13$ & $77.10 \pm 2.49$ & $80.88 \pm 1.36$ \\
 & TM-KMeans-Thermometer & $94.02 \pm 0.47$ & $76.79 \pm 1.69$ & $78.78 \pm 0.95$ \\
 & TM-MultiOtsu-Thermometer & $94.19 \pm 0.47$ & $75.61 \pm 2.19$ & $78.54 \pm 1.34$ \\
 & TM-Quantile-OneHot & $95.24 \pm 0.42$ & $73.73 \pm 3.52$ & $79.81 \pm 3.03$ \\
 & TM-Std & $94.48 \pm 0.60$ & $69.53 \pm 5.12$ & $75.29 \pm 4.94$ \\
\bottomrule
\end{tabular}
\end{table*}

\begin{table*}[htbp]
\centering
\caption{Comprehensive experimental results for the HBF1 dataset (Mean $\pm$ Standard Deviation across 5 CV folds).}
\label{tab:results_hbf1}
\begin{tabular}{llrrr}
\toprule
Task / Pair & Method & Test Acc & Test BAcc & Test F1 \\
\midrule
\multirow{17}{*}{Multiclass} & RF-Raw & $99.46 \pm 0.11$ & $95.60 \pm 0.54$ & $96.04 \pm 0.20$ \\
 & TM-SHB & $99.61 \pm 0.04$ & $95.60 \pm 0.44$ & $95.93 \pm 0.35$ \\
 & RF-SHB & $99.61 \pm 0.03$ & $95.30 \pm 0.38$ & $95.85 \pm 0.23$ \\
 & TM-KMeans-OneHot & $99.15 \pm 0.14$ & $95.13 \pm 0.46$ & $95.48 \pm 0.21$ \\
 & TM-KMeans-Thermometer & $99.01 \pm 0.38$ & $95.10 \pm 0.94$ & $95.34 \pm 0.82$ \\
 & TM-MultiOtsu-OneHot & $98.98 \pm 0.09$ & $94.86 \pm 0.55$ & $95.19 \pm 0.28$ \\
 & RF-KMeans-Thermometer & $99.31 \pm 0.04$ & $94.83 \pm 0.67$ & $95.52 \pm 0.32$ \\
 & RF-MultiOtsu-Thermometer & $99.19 \pm 0.04$ & $94.77 \pm 0.52$ & $95.35 \pm 0.17$ \\
 & RF-KMeans-OneHot & $99.33 \pm 0.02$ & $94.64 \pm 0.51$ & $95.43 \pm 0.24$ \\
 & TM-MultiOtsu-Thermometer & $98.94 \pm 0.45$ & $94.60 \pm 0.81$ & $95.09 \pm 0.56$ \\
 & RF-MultiOtsu-OneHot & $99.19 \pm 0.04$ & $94.47 \pm 0.46$ & $95.22 \pm 0.24$ \\
 & RF-Quantile-OneHot & $98.74 \pm 0.03$ & $93.93 \pm 0.39$ & $94.56 \pm 0.13$ \\
 & RF-Quantile-Thermometer & $98.74 \pm 0.03$ & $93.85 \pm 0.58$ & $94.53 \pm 0.22$ \\
 & RF-Std & $98.72 \pm 0.04$ & $93.72 \pm 0.76$ & $94.33 \pm 0.40$ \\
 & TM-Quantile-Thermometer & $98.11 \pm 0.45$ & $93.38 \pm 0.44$ & $93.82 \pm 0.52$ \\
 & TM-Quantile-OneHot & $98.52 \pm 0.07$ & $93.18 \pm 0.68$ & $94.18 \pm 0.24$ \\
 & TM-Std & $98.22 \pm 0.40$ & $91.63 \pm 0.59$ & $93.11 \pm 0.50$ \\
\midrule
\multirow{17}{*}{Blood-vs-Non-Blood} & TM-Quantile-OneHot & $99.62 \pm 0.04$ & $99.68 \pm 0.04$ & $99.56 \pm 0.05$ \\
 & TM-Std & $99.63 \pm 0.02$ & $99.67 \pm 0.03$ & $99.57 \pm 0.02$ \\
 & TM-SHB & $99.63 \pm 0.03$ & $99.66 \pm 0.05$ & $99.57 \pm 0.04$ \\
 & TM-Quantile-Thermometer & $99.60 \pm 0.03$ & $99.65 \pm 0.04$ & $99.54 \pm 0.04$ \\
 & RF-Quantile-Thermometer & $99.63 \pm 0.03$ & $99.65 \pm 0.03$ & $99.58 \pm 0.03$ \\
 & TM-MultiOtsu-Thermometer & $99.63 \pm 0.02$ & $99.65 \pm 0.02$ & $99.57 \pm 0.02$ \\
 & RF-Quantile-OneHot & $99.64 \pm 0.04$ & $99.65 \pm 0.03$ & $99.58 \pm 0.05$ \\
 & RF-KMeans-Thermometer & $99.67 \pm 0.02$ & $99.65 \pm 0.05$ & $99.62 \pm 0.03$ \\
 & RF-Std & $99.64 \pm 0.02$ & $99.65 \pm 0.03$ & $99.58 \pm 0.03$ \\
 & TM-KMeans-OneHot & $99.65 \pm 0.04$ & $99.64 \pm 0.05$ & $99.59 \pm 0.05$ \\
 & TM-KMeans-Thermometer & $99.63 \pm 0.04$ & $99.63 \pm 0.03$ & $99.58 \pm 0.05$ \\
 & RF-MultiOtsu-Thermometer & $99.65 \pm 0.02$ & $99.63 \pm 0.04$ & $99.60 \pm 0.03$ \\
 & RF-KMeans-OneHot & $99.65 \pm 0.02$ & $99.62 \pm 0.04$ & $99.59 \pm 0.02$ \\
 & TM-MultiOtsu-OneHot & $99.63 \pm 0.02$ & $99.61 \pm 0.05$ & $99.57 \pm 0.02$ \\
 & RF-Raw & $99.66 \pm 0.01$ & $99.61 \pm 0.04$ & $99.60 \pm 0.01$ \\
 & RF-MultiOtsu-OneHot & $99.63 \pm 0.02$ & $99.60 \pm 0.04$ & $99.58 \pm 0.02$ \\
 & RF-SHB & $99.64 \pm 0.05$ & $99.60 \pm 0.08$ & $99.58 \pm 0.05$ \\
\bottomrule
\end{tabular}
\end{table*}

\begin{table*}[htbp]
\centering
\caption{Comprehensive experimental results for the INPI dataset (Mean $\pm$ Standard Deviation across 5 CV folds). Ordered by BAcc.}
\label{tab:results_inpi}
\scriptsize
\begin{tabular}{llrrr}
\toprule
Task / Pair & Method & Test Acc & Test BAcc & Test F1 \\
\midrule
\multirow{17}{*}{Multiclass} & RF-SHB & $89.53 \pm 0.70$ & $86.50 \pm 3.74$ & $88.05 \pm 2.39$ \\
 & TM-SHB & $86.51 \pm 1.81$ & $78.96 \pm 3.42$ & $81.92 \pm 3.33$ \\
 & RF-Raw & $84.71 \pm 0.68$ & $77.96 \pm 1.47$ & $80.92 \pm 1.58$ \\
 & RF-MultiOtsu-OneHot & $76.67 \pm 0.63$ & $71.34 \pm 3.33$ & $72.82 \pm 3.35$ \\
 & RF-MultiOtsu-Thermometer & $76.44 \pm 0.68$ & $70.74 \pm 3.46$ & $72.59 \pm 3.06$ \\
 & RF-KMeans-OneHot & $76.03 \pm 1.06$ & $69.82 \pm 2.33$ & $72.48 \pm 2.22$ \\
 & RF-KMeans-Thermometer & $75.74 \pm 0.90$ & $69.18 \pm 1.78$ & $71.80 \pm 1.84$ \\
 & RF-Quantile-Thermometer & $76.86 \pm 0.52$ & $68.22 \pm 0.81$ & $69.91 \pm 1.11$ \\
 & RF-Std & $76.66 \pm 0.95$ & $67.75 \pm 1.09$ & $69.72 \pm 1.05$ \\
 & RF-Quantile-OneHot & $76.10 \pm 0.59$ & $67.25 \pm 1.00$ & $68.88 \pm 1.57$ \\
 & TM-KMeans-OneHot & $71.85 \pm 0.86$ & $62.42 \pm 1.21$ & $64.85 \pm 1.22$ \\
 & TM-MultiOtsu-OneHot & $71.87 \pm 0.86$ & $61.31 \pm 2.40$ & $64.36 \pm 2.53$ \\
 & TM-Quantile-OneHot & $71.70 \pm 1.04$ & $60.39 \pm 0.98$ & $62.03 \pm 1.78$ \\
 & TM-MultiOtsu-Thermometer & $70.30 \pm 0.55$ & $59.84 \pm 1.81$ & $61.74 \pm 1.65$ \\
 & TM-Std & $68.54 \pm 0.83$ & $59.78 \pm 2.88$ & $61.54 \pm 3.02$ \\
 & TM-Quantile-Thermometer & $71.04 \pm 1.28$ & $59.06 \pm 1.55$ & $59.64 \pm 2.07$ \\
 & TM-KMeans-Thermometer & $69.78 \pm 0.88$ & $58.87 \pm 0.52$ & $60.16 \pm 1.64$ \\
\midrule
\multirow{17}{*}{Corn-notill-vs-Soybean-notill} & TM-SHB & $97.36 \pm 0.44$ & $96.34 \pm 0.66$ & $96.50 \pm 0.59$ \\
 & RF-SHB & $96.85 \pm 0.39$ & $95.08 \pm 0.53$ & $95.77 \pm 0.52$ \\
 & RF-Raw & $92.20 \pm 0.42$ & $86.65 \pm 0.85$ & $88.98 \pm 0.64$ \\
 & TM-MultiOtsu-OneHot & $90.19 \pm 0.66$ & $83.79 \pm 0.70$ & $86.08 \pm 0.79$ \\
 & RF-MultiOtsu-Thermometer & $90.63 \pm 0.43$ & $83.76 \pm 1.29$ & $86.49 \pm 0.84$ \\
 & RF-Std & $90.44 \pm 0.76$ & $83.32 \pm 1.48$ & $86.15 \pm 1.23$ \\
 & TM-Quantile-Thermometer & $87.92 \pm 1.00$ & $83.27 \pm 2.26$ & $83.74 \pm 1.46$ \\
 & RF-MultiOtsu-OneHot & $90.15 \pm 0.68$ & $82.93 \pm 1.61$ & $85.72 \pm 1.22$ \\
 & RF-KMeans-Thermometer & $89.98 \pm 0.93$ & $82.56 \pm 2.17$ & $85.41 \pm 1.67$ \\
 & RF-KMeans-OneHot & $89.83 \pm 1.10$ & $82.44 \pm 2.29$ & $85.22 \pm 1.90$ \\
 & TM-Quantile-OneHot & $88.33 \pm 0.99$ & $80.96 \pm 2.16$ & $83.26 \pm 1.73$ \\
 & TM-KMeans-OneHot & $88.31 \pm 0.70$ & $79.55 \pm 1.97$ & $82.63 \pm 1.47$ \\
 & RF-Quantile-Thermometer & $87.53 \pm 0.73$ & $78.52 \pm 1.47$ & $81.51 \pm 1.31$ \\
 & RF-Quantile-OneHot & $87.34 \pm 0.60$ & $78.05 \pm 1.26$ & $81.11 \pm 1.03$ \\
 & TM-KMeans-Thermometer & $87.26 \pm 0.79$ & $77.74 \pm 1.50$ & $80.90 \pm 1.37$ \\
 & TM-MultiOtsu-Thermometer & $86.15 \pm 2.65$ & $77.09 \pm 3.66$ & $79.67 \pm 3.98$ \\
 & TM-Std & $87.34 \pm 1.00$ & $76.53 \pm 2.58$ & $80.26 \pm 2.24$ \\
\midrule
\multirow{17}{*}{Soybean-mintill-vs-Corn-mintill} & TM-SHB & $97.36 \pm 0.54$ & $96.46 \pm 0.68$ & $96.66 \pm 0.67$ \\
 & RF-SHB & $96.60 \pm 1.08$ & $95.30 \pm 1.71$ & $95.68 \pm 1.39$ \\
 & RF-Raw & $90.57 \pm 1.11$ & $87.06 \pm 1.65$ & $87.86 \pm 1.46$ \\
 & TM-Quantile-Thermometer & $88.58 \pm 0.97$ & $86.98 \pm 1.44$ & $86.00 \pm 1.04$ \\
 & RF-Std & $89.72 \pm 2.48$ & $86.90 \pm 3.12$ & $87.02 \pm 3.06$ \\
 & RF-MultiOtsu-Thermometer & $89.53 \pm 2.04$ & $86.67 \pm 2.87$ & $86.77 \pm 2.56$ \\
 & TM-KMeans-OneHot & $88.58 \pm 2.37$ & $85.91 \pm 3.08$ & $85.71 \pm 2.93$ \\
 & RF-KMeans-Thermometer & $88.49 \pm 0.86$ & $85.41 \pm 1.78$ & $85.47 \pm 1.11$ \\
 & TM-MultiOtsu-Thermometer & $89.25 \pm 1.61$ & $85.40 \pm 1.13$ & $86.19 \pm 1.78$ \\
 & RF-MultiOtsu-OneHot & $88.77 \pm 1.87$ & $85.07 \pm 2.98$ & $85.61 \pm 2.55$ \\
 & RF-Quantile-Thermometer & $87.55 \pm 2.23$ & $84.66 \pm 2.84$ & $84.43 \pm 2.77$ \\
 & TM-MultiOtsu-OneHot & $87.92 \pm 1.69$ & $84.59 \pm 1.33$ & $84.76 \pm 1.86$ \\
 & TM-Quantile-OneHot & $88.68 \pm 1.60$ & $84.58 \pm 1.74$ & $85.41 \pm 1.78$ \\
 & TM-KMeans-Thermometer & $87.92 \pm 2.42$ & $84.49 \pm 3.49$ & $84.68 \pm 3.15$ \\
 & TM-Std & $88.77 \pm 1.47$ & $83.89 \pm 2.04$ & $85.26 \pm 1.96$ \\
 & RF-Quantile-OneHot & $87.36 \pm 2.30$ & $83.67 \pm 3.41$ & $83.93 \pm 3.06$ \\
 & RF-KMeans-OneHot & $87.45 \pm 2.25$ & $83.41 \pm 2.36$ & $83.97 \pm 2.61$ \\
\midrule
\multirow{17}{*}{Woods-vs-Grass-pasture} & TM-SHB & $95.96 \pm 0.89$ & $95.29 \pm 1.11$ & $95.46 \pm 1.00$ \\
 & RF-SHB & $94.11 \pm 0.96$ & $92.92 \pm 1.51$ & $93.34 \pm 1.11$ \\
 & TM-Quantile-Thermometer & $93.97 \pm 1.89$ & $92.35 \pm 2.72$ & $93.10 \pm 2.24$ \\
 & TM-Std & $93.97 \pm 0.83$ & $91.78 \pm 0.74$ & $93.04 \pm 0.93$ \\
 & TM-Quantile-OneHot & $92.70 \pm 1.78$ & $90.76 \pm 2.58$ & $91.61 \pm 2.17$ \\
 & TM-MultiOtsu-OneHot & $92.84 \pm 1.21$ & $90.25 \pm 1.93$ & $91.66 \pm 1.51$ \\
 & RF-Raw & $92.34 \pm 2.09$ & $90.13 \pm 2.72$ & $91.17 \pm 2.49$ \\
 & RF-MultiOtsu-Thermometer & $92.48 \pm 1.45$ & $90.03 \pm 2.06$ & $91.28 \pm 1.73$ \\
 & TM-KMeans-OneHot & $92.70 \pm 1.08$ & $89.88 \pm 1.61$ & $91.45 \pm 1.33$ \\
 & RF-Std & $92.55 \pm 1.48$ & $89.88 \pm 1.93$ & $91.32 \pm 1.76$ \\
 & RF-Quantile-Thermometer & $91.99 \pm 1.34$ & $89.81 \pm 2.00$ & $90.78 \pm 1.61$ \\
 & TM-MultiOtsu-Thermometer & $92.41 \pm 0.40$ & $89.62 \pm 0.78$ & $91.14 \pm 0.54$ \\
 & RF-KMeans-Thermometer & $91.91 \pm 1.73$ & $88.88 \pm 2.48$ & $90.50 \pm 2.12$ \\
 & TM-KMeans-Thermometer & $91.84 \pm 2.02$ & $88.52 \pm 2.60$ & $90.36 \pm 2.42$ \\
 & RF-MultiOtsu-OneHot & $91.13 \pm 1.23$ & $88.08 \pm 1.72$ & $89.62 \pm 1.49$ \\
 & RF-KMeans-OneHot & $91.06 \pm 2.00$ & $87.72 \pm 2.79$ & $89.44 \pm 2.47$ \\
 & RF-Quantile-OneHot & $89.50 \pm 1.75$ & $86.13 \pm 2.47$ & $87.66 \pm 2.19$ \\
\bottomrule
\end{tabular}
\end{table*}

\begin{table*}[htbp]
\centering
\caption{Comprehensive experimental results for the MOSA dataset (Mean $\pm$ Standard Deviation across 5 CV folds). Ordered by BAcc.}
\label{tab:results_mosa}
\begin{tabular}{llrrr}
\toprule
Task / Pair & Method & Test Acc & Test BAcc & Test F1 \\
\midrule
\multirow{17}{*}{Multiclass} & RF-SHB & $83.90 \pm 0.69$ & $77.05 \pm 0.91$ & $80.13 \pm 0.90$ \\
 & TM-SHB & $77.28 \pm 0.92$ & $65.27 \pm 1.40$ & $66.90 \pm 5.06$ \\
 & RF-Raw & $75.09 \pm 0.45$ & $63.81 \pm 0.51$ & $67.46 \pm 0.62$ \\
 & RF-MultiOtsu-Thermometer & $68.80 \pm 0.55$ & $55.51 \pm 0.54$ & $59.39 \pm 0.57$ \\
 & RF-MultiOtsu-OneHot & $68.77 \pm 0.56$ & $55.48 \pm 0.57$ & $59.37 \pm 0.66$ \\
 & RF-Std & $68.38 \pm 0.54$ & $55.48 \pm 0.85$ & $58.76 \pm 1.00$ \\
 & RF-KMeans-Thermometer & $68.39 \pm 0.55$ & $55.04 \pm 0.49$ & $59.03 \pm 0.58$ \\
 & RF-KMeans-OneHot & $68.45 \pm 0.48$ & $55.00 \pm 0.45$ & $59.03 \pm 0.51$ \\
 & RF-Quantile-Thermometer & $68.10 \pm 0.65$ & $54.56 \pm 0.80$ & $58.58 \pm 0.80$ \\
 & RF-Quantile-OneHot & $68.02 \pm 0.65$ & $54.30 \pm 0.77$ & $58.44 \pm 0.74$ \\
 & TM-KMeans-OneHot & $60.68 \pm 0.77$ & $43.01 \pm 0.92$ & $44.08 \pm 2.98$ \\
 & TM-MultiOtsu-OneHot & $60.42 \pm 0.84$ & $42.02 \pm 1.02$ & $42.06 \pm 2.38$ \\
 & TM-Quantile-OneHot & $59.37 \pm 1.52$ & $41.55 \pm 1.69$ & $43.55 \pm 1.93$ \\
 & TM-MultiOtsu-Thermometer & $57.76 \pm 2.08$ & $41.15 \pm 1.11$ & $42.88 \pm 0.85$ \\
 & TM-KMeans-Thermometer & $59.31 \pm 0.47$ & $40.99 \pm 0.32$ & $41.74 \pm 2.16$ \\
 & TM-Std & $58.79 \pm 0.53$ & $40.26 \pm 0.55$ & $39.45 \pm 3.10$ \\
 & TM-Quantile-Thermometer & $58.89 \pm 1.63$ & $39.93 \pm 1.56$ & $41.14 \pm 1.70$ \\
\midrule
\multirow{17}{*}{BAEL-vs-DIPA} & RF-SHB & $94.13 \pm 0.34$ & $88.01 \pm 0.63$ & $90.29 \pm 0.58$ \\
 & TM-SHB & $92.78 \pm 0.21$ & $84.66 \pm 1.17$ & $87.70 \pm 0.60$ \\
 & RF-Raw & $89.66 \pm 0.35$ & $79.15 \pm 0.61$ & $82.14 \pm 0.60$ \\
 & RF-MultiOtsu-Thermometer & $87.78 \pm 0.43$ & $75.94 \pm 0.77$ & $78.74 \pm 0.79$ \\
 & RF-Std & $87.47 \pm 0.24$ & $75.75 \pm 0.53$ & $78.35 \pm 0.48$ \\
 & RF-KMeans-Thermometer & $87.65 \pm 0.18$ & $75.70 \pm 0.31$ & $78.49 \pm 0.29$ \\
 & RF-KMeans-OneHot & $87.74 \pm 0.18$ & $75.67 \pm 0.34$ & $78.55 \pm 0.30$ \\
 & RF-MultiOtsu-OneHot & $87.73 \pm 0.18$ & $75.65 \pm 0.48$ & $78.54 \pm 0.39$ \\
 & RF-Quantile-Thermometer & $87.69 \pm 0.31$ & $75.48 \pm 0.74$ & $78.40 \pm 0.67$ \\
 & RF-Quantile-OneHot & $87.53 \pm 0.32$ & $74.81 \pm 0.69$ & $77.90 \pm 0.61$ \\
 & TM-Quantile-OneHot & $84.68 \pm 0.39$ & $69.87 \pm 1.72$ & $72.44 \pm 1.39$ \\
 & TM-MultiOtsu-OneHot & $83.66 \pm 1.33$ & $68.75 \pm 1.75$ & $70.95 \pm 1.40$ \\
 & TM-Quantile-Thermometer & $82.25 \pm 1.08$ & $67.57 \pm 1.37$ & $69.22 \pm 0.68$ \\
 & TM-Std & $83.93 \pm 0.19$ & $66.94 \pm 1.97$ & $69.68 \pm 1.87$ \\
 & TM-KMeans-OneHot & $83.23 \pm 1.82$ & $66.47 \pm 1.93$ & $68.91 \pm 2.17$ \\
 & TM-MultiOtsu-Thermometer & $83.05 \pm 0.69$ & $63.67 \pm 1.92$ & $66.24 \pm 2.03$ \\
 & TM-KMeans-Thermometer & $82.73 \pm 0.81$ & $63.19 \pm 2.30$ & $65.61 \pm 2.71$ \\
\midrule
\multirow{17}{*}{LEAM-vs-CEPE} & TM-SHB & $96.18 \pm 0.24$ & $96.18 \pm 0.24$ & $96.18 \pm 0.24$ \\
 & RF-SHB & $95.84 \pm 0.13$ & $95.83 \pm 0.13$ & $95.84 \pm 0.13$ \\
 & RF-Raw & $91.93 \pm 0.41$ & $91.94 \pm 0.40$ & $91.93 \pm 0.41$ \\
 & RF-Quantile-Thermometer & $89.05 \pm 0.25$ & $89.05 \pm 0.25$ & $89.05 \pm 0.25$ \\
 & RF-Quantile-OneHot & $88.92 \pm 0.50$ & $88.92 \pm 0.50$ & $88.92 \pm 0.50$ \\
 & RF-MultiOtsu-Thermometer & $88.85 \pm 0.72$ & $88.85 \pm 0.73$ & $88.85 \pm 0.72$ \\
 & RF-MultiOtsu-OneHot & $88.72 \pm 0.39$ & $88.72 \pm 0.38$ & $88.72 \pm 0.38$ \\
 & RF-KMeans-Thermometer & $88.60 \pm 0.48$ & $88.59 \pm 0.48$ & $88.59 \pm 0.48$ \\
 & RF-Std & $88.44 \pm 0.55$ & $88.45 \pm 0.54$ & $88.43 \pm 0.55$ \\
 & TM-MultiOtsu-OneHot & $88.21 \pm 0.38$ & $88.21 \pm 0.39$ & $88.21 \pm 0.38$ \\
 & TM-Quantile-Thermometer & $87.89 \pm 0.60$ & $87.89 \pm 0.60$ & $87.89 \pm 0.60$ \\
 & RF-KMeans-OneHot & $87.72 \pm 0.73$ & $87.72 \pm 0.73$ & $87.72 \pm 0.73$ \\
 & TM-MultiOtsu-Thermometer & $87.67 \pm 0.96$ & $87.67 \pm 0.97$ & $87.67 \pm 0.97$ \\
 & TM-Std & $87.45 \pm 0.51$ & $87.45 \pm 0.51$ & $87.44 \pm 0.51$ \\
 & TM-KMeans-Thermometer & $87.40 \pm 0.65$ & $87.40 \pm 0.65$ & $87.40 \pm 0.65$ \\
 & TM-KMeans-OneHot & $87.17 \pm 0.51$ & $87.16 \pm 0.52$ & $87.16 \pm 0.51$ \\
 & TM-Quantile-OneHot & $87.11 \pm 0.74$ & $87.11 \pm 0.74$ & $87.10 \pm 0.74$ \\
\bottomrule
\end{tabular}
\end{table*}

\begin{table*}[htbp]
\centering
\caption{Comprehensive experimental results for the PAVI dataset (Mean $\pm$ Standard Deviation across 5 CV folds). Ordered by BAcc.}
\label{tab:results_pavi}
\begin{tabular}{llrrr}
\toprule
Task / Pair & Method & Test Acc & Test BAcc & Test F1 \\
\midrule
\multirow{17}{*}{Multiclass} & RF-SHB & $93.77 \pm 0.30$ & $90.88 \pm 0.42$ & $91.90 \pm 0.41$ \\
 & RF-Raw & $92.71 \pm 0.26$ & $90.32 \pm 0.33$ & $91.39 \pm 0.27$ \\
 & TM-SHB & $92.92 \pm 0.30$ & $88.66 \pm 0.39$ & $90.49 \pm 0.43$ \\
 & RF-Quantile-OneHot & $85.25 \pm 0.25$ & $80.25 \pm 0.30$ & $81.23 \pm 0.25$ \\
 & RF-Quantile-Thermometer & $85.23 \pm 0.15$ & $80.11 \pm 0.20$ & $81.11 \pm 0.16$ \\
 & RF-Std & $83.56 \pm 0.35$ & $79.53 \pm 0.49$ & $80.61 \pm 0.55$ \\
 & RF-MultiOtsu-OneHot & $83.60 \pm 0.29$ & $78.69 \pm 0.38$ & $79.69 \pm 0.46$ \\
 & RF-MultiOtsu-Thermometer & $83.57 \pm 0.38$ & $78.68 \pm 0.42$ & $79.68 \pm 0.53$ \\
 & RF-KMeans-OneHot & $81.85 \pm 0.55$ & $75.79 \pm 1.03$ & $76.69 \pm 0.94$ \\
 & RF-KMeans-Thermometer & $81.86 \pm 0.39$ & $75.75 \pm 0.96$ & $76.67 \pm 0.82$ \\
 & TM-Quantile-OneHot & $83.35 \pm 0.59$ & $75.38 \pm 0.94$ & $75.99 \pm 3.80$ \\
 & TM-Quantile-Thermometer & $82.69 \pm 0.64$ & $74.88 \pm 0.72$ & $73.76 \pm 4.06$ \\
 & TM-MultiOtsu-Thermometer & $80.43 \pm 0.27$ & $73.42 \pm 0.51$ & $67.90 \pm 0.50$ \\
 & TM-MultiOtsu-OneHot & $80.91 \pm 0.61$ & $72.85 \pm 1.69$ & $73.60 \pm 3.02$ \\
 & TM-Std & $80.40 \pm 0.45$ & $71.95 \pm 0.76$ & $70.25 \pm 3.26$ \\
 & TM-KMeans-Thermometer & $77.72 \pm 1.04$ & $70.10 \pm 1.79$ & $70.56 \pm 1.73$ \\
 & TM-KMeans-OneHot & $78.93 \pm 0.88$ & $68.81 \pm 1.41$ & $70.54 \pm 2.10$ \\
\midrule
\multirow{17}{*}{Asphalt-vs-Bitumen} & TM-SHB & $96.20 \pm 0.03$ & $90.40 \pm 0.73$ & $92.77 \pm 0.14$ \\
 & RF-Raw & $95.27 \pm 0.67$ & $90.27 \pm 1.38$ & $91.29 \pm 1.23$ \\
 & RF-SHB & $96.11 \pm 0.52$ & $90.01 \pm 1.35$ & $92.56 \pm 1.05$ \\
 & RF-KMeans-Thermometer & $93.58 \pm 0.56$ & $87.87 \pm 1.48$ & $88.33 \pm 1.02$ \\
 & RF-KMeans-OneHot & $93.75 \pm 0.66$ & $87.73 \pm 1.85$ & $88.53 \pm 1.35$ \\
 & RF-Std & $93.79 \pm 0.69$ & $87.56 \pm 1.30$ & $88.56 \pm 1.17$ \\
 & TM-Quantile-OneHot & $92.81 \pm 0.87$ & $86.68 \pm 1.98$ & $86.98 \pm 1.44$ \\
 & RF-MultiOtsu-Thermometer & $93.56 \pm 0.51$ & $86.68 \pm 1.56$ & $88.00 \pm 1.05$ \\
 & RF-MultiOtsu-OneHot & $93.27 \pm 0.38$ & $86.12 \pm 1.43$ & $87.46 \pm 0.65$ \\
 & TM-Quantile-Thermometer & $87.33 \pm 3.35$ & $85.82 \pm 0.71$ & $80.66 \pm 3.54$ \\
 & RF-Quantile-Thermometer & $93.46 \pm 0.26$ & $85.63 \pm 1.27$ & $87.59 \pm 0.69$ \\
 & RF-Quantile-OneHot & $93.17 \pm 0.50$ & $84.43 \pm 2.13$ & $86.82 \pm 1.28$ \\
 & TM-KMeans-OneHot & $93.37 \pm 1.13$ & $84.19 \pm 4.30$ & $86.94 \pm 2.84$ \\
 & TM-KMeans-Thermometer & $91.19 \pm 1.13$ & $83.46 \pm 0.84$ & $83.98 \pm 1.58$ \\
 & TM-MultiOtsu-OneHot & $93.00 \pm 0.89$ & $82.93 \pm 3.33$ & $86.07 \pm 2.26$ \\
 & TM-Std & $91.74 \pm 0.57$ & $82.23 \pm 1.75$ & $84.21 \pm 1.31$ \\
 & TM-MultiOtsu-Thermometer & $90.28 \pm 0.52$ & $82.02 \pm 1.33$ & $82.35 \pm 0.92$ \\
\midrule
\multirow{17}{*}{Meadows-vs-Trees} & TM-SHB & $99.10 \pm 0.10$ & $97.89 \pm 0.22$ & $98.14 \pm 0.20$ \\
 & RF-SHB & $98.92 \pm 0.15$ & $97.38 \pm 0.24$ & $97.75 \pm 0.31$ \\
 & RF-Raw & $98.07 \pm 0.08$ & $95.38 \pm 0.46$ & $95.97 \pm 0.18$ \\
 & TM-Std & $97.82 \pm 0.15$ & $94.53 \pm 0.74$ & $95.42 \pm 0.36$ \\
 & RF-Std & $97.80 \pm 0.10$ & $94.37 \pm 0.18$ & $95.36 \pm 0.21$ \\
 & TM-MultiOtsu-Thermometer & $97.62 \pm 0.07$ & $94.20 \pm 0.50$ & $95.01 \pm 0.13$ \\
 & RF-MultiOtsu-Thermometer & $97.62 \pm 0.16$ & $94.07 \pm 0.42$ & $94.99 \pm 0.34$ \\
 & TM-MultiOtsu-OneHot & $97.57 \pm 0.16$ & $94.02 \pm 0.57$ & $94.89 \pm 0.34$ \\
 & RF-MultiOtsu-OneHot & $97.55 \pm 0.13$ & $93.88 \pm 0.42$ & $94.84 \pm 0.28$ \\
 & TM-Quantile-Thermometer & $96.48 \pm 0.70$ & $93.84 \pm 0.57$ & $92.92 \pm 1.25$ \\
 & RF-Quantile-OneHot & $97.05 \pm 0.22$ & $93.61 \pm 0.40$ & $93.87 \pm 0.45$ \\
 & RF-Quantile-Thermometer & $97.10 \pm 0.10$ & $93.42 \pm 0.42$ & $93.95 \pm 0.24$ \\
 & TM-Quantile-OneHot & $96.92 \pm 0.18$ & $92.98 \pm 1.02$ & $93.56 \pm 0.42$ \\
 & RF-KMeans-Thermometer & $96.62 \pm 0.40$ & $92.08 \pm 0.52$ & $92.90 \pm 0.79$ \\
 & RF-KMeans-OneHot & $96.56 \pm 0.42$ & $91.95 \pm 0.59$ & $92.78 \pm 0.83$ \\
 & TM-KMeans-Thermometer & $96.59 \pm 0.28$ & $91.71 \pm 1.10$ & $92.78 \pm 0.45$ \\
 & TM-KMeans-OneHot & $96.15 \pm 0.60$ & $89.03 \pm 1.86$ & $91.53 \pm 1.35$ \\
\bottomrule
\end{tabular}
\end{table*}

\clearpage

\bibliographystyle{cas-model2-names}
\bibliography{stm_bibliography}

@article{Li2019deep,
  author  = {Li, Shutao and Song, Weiwei and Fang, Leyuan and Chen, Yushi and Ghamisi, Pedram and Benediktsson, J{\'o}n Atli},
  title   = {Deep Learning for Hyperspectral Image Classification: An Overview},
  journal = {IEEE Transactions on Geoscience and Remote Sensing},
  year    = {2019},  volume  = {57},  number  = {9},  pages   = {6690--6709},  doi     = {10.1109/TGRS.2019.2907932}}

@article{Rasti2020feature,
  author  = {Rasti, Behnood and Hong, Danfeng and Hang, Renlong and Ghamisi, Pedram and Kang, Xudong and Chanussot, Jocelyn and Benediktsson, J{\'o}n Atli},
  title   = {Feature Extraction for Hyperspectral Imagery: The Evolution from Shallow to Deep: Overview and Toolbox},
  journal = {IEEE Geoscience and Remote Sensing Magazine},
  year    = {2020},  volume  = {8},  number  = {4},  pages   = {60--88},  doi     = {10.1109/MGRS.2020.2979764}}

@inproceedings{Joelsson2005random,
  author    = {Joelsson, Sveinn R. and Benediktsson, J{\'o}n Atli and Sveinsson, Johannes R.},
  title     = {Random Forest Classifiers for Hyperspectral Data},
  booktitle = {2005 IEEE International Geoscience and Remote Sensing Symposium},
  year      = {2005},  doi       = {10.1109/IGARSS.2005.1526129}}

@article{Pal2005random,
  author  = {Pal, Mahesh},
  title   = {Random Forest Classifier for Remote Sensing Classification},
  journal = {International Journal of Remote Sensing},
  year    = {2005},  volume  = {26},  number  = {1},  pages   = {217--222},  doi     = {10.1080/01431160412331269698}}

@article{Breiman2001random,
  author  = {Breiman, Leo},
  title   = {Random Forests},
  journal = {Machine Learning},
  year    = {2001},  volume  = {45},  number  = {1},  pages   = {5--32},  doi     = {10.1023/A:1010933404324}}

@article{Viola2004robust,
  author  = {Viola, Paul and Jones, Michael J.},
  title   = {Robust Real-Time Face Detection},
  journal = {International Journal of Computer Vision},
  year    = {2004},  volume  = {57},  number  = {2},  pages   = {137--154},  doi     = {10.1023/B:VISI.0000013087.49260.fb}}

@article{Lloyd1982least,
  author  = {Lloyd, Stuart P.},
  title   = {Least Squares Quantization in PCM},
  journal = {IEEE Transactions on Information Theory},
  year    = {1982},  volume  = {28},  number  = {2},  pages   = {129--137},  doi     = {10.1109/TIT.1982.1056489}}

@article{Liao2001fast,
  author  = {Liao, Ping-Sung and Chen, Tse-Sheng and Chung, Pau-Choo},
  title   = {A Fast Algorithm for Multilevel Thresholding},
  journal = {Journal of Information Science and Engineering},
  year    = {2001},  volume  = {17},  number  = {5},  pages   = {713--727}, doi = {10.6688/JISE.2001.17.5.1}}

@article{Otsu1979threshold,
  author  = {Otsu, Nobuyuki},
  title   = {A Threshold Selection Method from Gray-Level Histograms},
  journal = {IEEE Transactions on Systems, Man, and Cybernetics},
  year    = {1979},  volume  = {9},  number  = {1},  pages   = {62--66},  doi     = {10.1109/TSMC.1979.4310076}}

@article{Hancock2020categorical,
  author  = {Hancock, John T. and Khoshgoftaar, Taghi M.},
  title   = {Survey on Categorical Data for Neural Networks},
  journal = {Journal of Big Data},
  year    = {2020},  volume  = {7},  number  = {1},  pages   = {28},  doi     = {10.1186/s40537-020-00305-w}}

@article{Garcia2013survey,
  author  = {Garc{\'i}a, Salvador and Luengo, Juli{\'a}n and S{\'a}ez, Jos{\'e} Antonio and L{\'o}pez, Victoria and Herrera, Francisco},
  title   = {A Survey of Discretization Techniques: Taxonomy and Empirical Analysis in Supervised Learning},
  journal = {IEEE Transactions on Knowledge and Data Engineering},
  year    = {2013},  volume  = {25},  number  = {4},  pages   = {734--750},  doi     = {10.1109/TKDE.2012.35}}

@article{Romaszewski2021blood,
  author  = {Romaszewski, Micha{\l} and G{\l}omb, Przemys{\l}aw and Sochan, Arkadiusz and Cholewa, Micha{\l}},
  title   = {A Dataset for Evaluating Blood Detection in Hyperspectral Images},
  journal = {Forensic Science International},
  year    = {2021},  volume  = {320},  pages   = {110701},  doi     = {10.1016/j.forsciint.2021.110701}}

@article{Melgani2004svm,
  author  = {Melgani, Farid and Bruzzone, Lorenzo},
  title   = {Classification of Hyperspectral Remote Sensing Images with Support Vector Machines},
  journal = {IEEE Transactions on Geoscience and Remote Sensing},
  year    = {2004},  volume  = {42},  number  = {8},  pages   = {1778--1790},  doi     = {10.1109/TGRS.2004.831865}}

@article{Gonzales2013fpga,
title = {Use of FPGA or GPU-based architectures for remotely sensed hyperspectral image processing},
journal = {Integration},
volume = {46},number = {2},pages = {89-103},year = {2013},issn = {0167-9260},
doi = {10.1016/j.vlsi.2012.04.002},
author = {Carlos González and Sergio Sánchez and Abel Paz and Javier Resano and Daniel Mozos and Antonio Plaza}}

@Article{Yang2025onboard,
AUTHOR = {Yang, Ruifan and Huang, Min and Zhao, Wenhao and Zhang, Zixuan and Sun, Yan and Qian, Lulu and Wang, Zhanchao},
TITLE = {Onboard Real-Time Hyperspectral Image Processing System Design for Unmanned Aerial Vehicles},
JOURNAL = {Sensors},
VOLUME = {25},YEAR = {2025},NUMBER = {15},ARTICLE-NUMBER = {4822},
ISSN = {1424-8220},DOI = {10.3390/s25154822}}

@Article{Wan2025hsi,
author={Wan, Xiaoqing
and Chen, Feng
and Gao, Weizhe
and Mo, Dongtao
and Liu, Hui},
title={Fusion of circulant singular spectrum analysis and multiscale local ternary patterns for effective spectral-spatial feature extraction and small sample hyperspectral image classification},
journal={Scientific Reports},
year={2025},volume={15},number={1},pages={6972},
issn={2045-2322},
doi={10.1038/s41598-025-90926-z}}

@article{Ram2024agri,
title = {A systematic review of hyperspectral imaging in precision agriculture: Analysis of its current state and future prospects},
journal = {Computers and Electronics in Agriculture},
volume = {222},pages = {109037},year = {2024},issn = {0168-1699},
doi = {10.1016/j.compag.2024.109037},
author = {Billy G. Ram and Peter Oduor and C. Igathinathane and Kirk Howatt and Xin Sun}}

@article{Bhargava2024review,
title = {Hyperspectral imaging and its applications: A review},
journal = {Heliyon},
volume = {10},number = {12},pages = {e33208},year = {2024},issn = {2405-8440},
doi = {10.1016/j.heliyon.2024.e33208},
author = {Anuja Bhargava and Ashish Sachdeva and Kulbhushan Sharma and Mohammed H. Alsharif and Peerapong Uthansakul and Monthippa Uthansakul}}

@Article{Grewal2023survey,
AUTHOR = {Grewal, Reaya and Singh Kasana, Singara and Kasana, Geeta},
TITLE = {Machine Learning and Deep Learning Techniques for Spectral Spatial Classification of Hyperspectral Images: A Comprehensive Survey},
JOURNAL = {Electronics},
VOLUME = {12},YEAR = {2023},NUMBER = {3},ARTICLE-NUMBER = {488},
ISSN = {2079-9292},DOI = {10.3390/electronics12030488}}

@Article{Langer2023robust,
AUTHOR = {Langer, Dennis D. and Orlandić, Milica and Bakken, Sivert and Birkeland, Roger and Garrett, Joseph L. and Johansen, Tor A. and Sørensen, Asgeir J.},
TITLE = {Robust and Reconfigurable On-Board Processing for a Hyperspectral Imaging Small Satellite},
JOURNAL = {Remote Sensing},
VOLUME = {15},YEAR = {2023},NUMBER = {15},ARTICLE-NUMBER = {3756},
ISSN = {2072-4292},DOI = {10.3390/rs15153756}}

@article{Wheeldon2020automata,
    author = {Wheeldon, Adrian and Shafik, Rishad and Rahman, Tousif and Lei, Jie and Yakovlev, Alex and Granmo, Ole-Christoffer},
    title = {Learning automata based energy-efficient {AI} hardware design for {IoT} applications},
    journal = {Philosophical Transactions of the Royal Society A: Mathematical, Physical and Engineering Sciences},
    volume = {378},    number = {2182},    pages = {20190593},    year = {2020},
    issn = {1364-503X},    doi = {10.1098/rsta.2019.0593}}

@ARTICLE{Mao2025dynamic,
  author={Mao, Gang and Rahman, Tousif and Maheshwari, Sidharth and Pattison, Bob and Shao, Zhuang and Shafik, Rishad and Yakovlev, Alex},
  journal={IEEE Transactions on Circuits and Systems I: Regular Papers}, 
  title={Dynamic Tsetlin Machine Accelerators for On-Chip Training Using FPGAs}, 
  year={2025},   volume={72},   number={11},   pages={6962-6975},
  doi={10.1109/TCSI.2025.3564875}}

@ARTICLE{Tunheim2025fpga,
  author={Tunheim, Svein Anders and Jiao, Lei and Shafik, Rishad and Yakovlev, Alex and Granmo, Ole-Christoffer},
  journal={IEEE Transactions on Circuits and Systems I: Regular Papers}, 
  title={Tsetlin {M}achine-Based Image Classification {FPGA} Accelerator With On-Device Training}, 
  year={2025},  volume={72},  number={2},  pages={830-843},  doi={10.1109/TCSI.2024.3519191}}

@article{Saha2023discover,
author = {Saha, Rupsa and Granmo, Ole-Christoffer and Goodwin, Morten},
title = {Using {T}setlin {M}achine to discover interpretable rules in natural language processing applications},
journal = {Expert Systems},
volume = {40},number = {4},pages = {e12873},
doi = {10.1111/exsy.12873},year = {2023}}

@inproceedings{Abeyrathna2024concise,
author = {Abeyrathna, K. Darshana and Abouzeid, Ahmed A. O. and Bhattarai, Bimal and Giri, Charul and Glimsdal, Sondre and Granmo, Ole-Christoffer and Jiao, Lei and Saha, Rupsa and Sharma, Jivitesh and Tunheim, Svein A. and Zhang, Xuan},
title = {Building concise logical patterns by constraining tsetlin machine clause size},
year = {2023}, isbn = {978-1-956792-03-4},
doi = {10.24963/ijcai.2023/378},
booktitle = {Proceedings of the Thirty-Second International Joint Conference on Artificial Intelligence},
articleno = {378},numpages = {9},series = {IJCAI '23}}

@Article{Rudin2019stop,
author={Rudin, Cynthia},
title={Stop explaining black box machine learning models for high stakes decisions and use interpretable models instead},
journal={Nature Machine Intelligence},
year={2019},volume={1},number={5},pages={206-215},issn={2522-5839},doi={10.1038/s42256-019-0048-x}}

@Article{Leite2025fault,
AUTHOR = {Leite, Denis and Andrade, Emmanuel and Rativa, Diego and Maciel, Alexandre M. A.},
TITLE = {Fault Detection and Diagnosis in {Industry 4.0}: A Review on Challenges and Opportunities},
JOURNAL = {Sensors},VOLUME = {25},YEAR = {2025},NUMBER = {1},ARTICLE-NUMBER = {60},
ISSN = {1424-8220},DOI = {10.3390/s25010060}}

@Article{Soroush2025tiny,
AUTHOR = {Heydari, Soroush and Mahmoud, Qusay H.},
TITLE = {Tiny {M}achine {L}earning and On-Device Inference: A Survey of Applications, Challenges, and Future Directions},
JOURNAL = {Sensors},VOLUME = {25},YEAR = {2025},NUMBER = {10},ARTICLE-NUMBER = {3191},
ISSN = {1424-8220},DOI = {10.3390/s25103191}}

@article{Singh2023edge,
title = {Edge {AI}: A survey},
journal = {Internet of Things and Cyber-Physical Systems},
volume = {3},pages = {71-92},year = {2023},issn = {2667-3452},
doi = {10.1016/j.iotcps.2023.02.004},
author = {Raghubir Singh and Sukhpal Singh Gill}}

@article{Bommasani2022opportunities,
      title={On the Opportunities and Risks of Foundation Models}, 
      author={Rishi Bommasani and Drew A. Hudson and Ehsan Adeli and others},
      year={2022},       journal={arXiv 2108.07258}}

@article{Jordan2015trends,
author = {M. I. Jordan  and T. M. Mitchell },
title = {Machine learning: Trends, perspectives, and prospects},
journal = {Science},
volume = {349},number = {6245},pages = {255-260},year = {2015},doi = {10.1126/science.aaa8415}}

@inproceedings{viola2001rapid,
  title={Rapid object detection using a boosted cascade of simple features},
  author={Viola, Paul and Jones, Michael},
  booktitle={Proceedings of the 2001 IEEE computer society conference on computer vision and pattern recognition. CVPR 2001},
  volume={1},
  pages={I--I},
  year={2001},
  organization={Ieee}
}

@article{benavoli2017time,
  title={Time for a change: a tutorial for comparing multiple classifiers through Bayesian analysis},
  author={Benavoli, Alessio and Corani, Giorgio and Dem{\v{s}}ar, Janez and Zaffalon, Marco},
  journal={Journal of Machine Learning Research},
  volume={18},
  number={77},
  pages={1--36},
  year={2017}
}

@inproceedings{rahimi2019tsetlin,
  title={A Tsetlin machine with multigranular clauses},
  author={Rahimi Gorji, Saeed and Granmo, Ole-Christoffer and Phoulady, Adrian and Goodwin, Morten},
  booktitle={International Conference on Innovative Techniques and Applications of Artificial Intelligence},
  pages={146--151},
  year={2019},
  organization={Springer}
}

@article{clark2005hyperspectral,
  title={Hyperspectral discrimination of tropical rain forest tree species at leaf to crown scales},
  author={Clark, Matthew L and Roberts, Dar A and Clark, David B},
  journal={Remote sensing of environment},
  volume={96},
  number={3-4},
  pages={375--398},
  year={2005},
  publisher={Elsevier}
}

@article{granmo2018tsetlin,
  author = {{Granmo}, Ole-Christoffer},
  title = "{The Tsetlin Machine - A Game Theoretic Bandit Driven Approach to Optimal Pattern Recognition with Propositional Logic}",
  journal = {arXiv preprint arXiv:1804.01508}, year = {2018},
  url={https://arxiv.org/abs/1804.01508}
}

@article{abeyrathna2019continuous,
  title={A Scheme for Continuous Input to the Tsetlin Machine with Applications to Forecasting Disease Outbreaks},
  author={Abeyrathna, K. Darshana and Granmo, Ole-Christoffer and Zhang, Xuan and Goodwin, Morten},
  journal={arXiv preprint arXiv:1905.04199},
  year={2019}
}

@article{granmo2019convolutional,
  title={The Convolutional Tsetlin Machine},
  author={Granmo, Ole-Christoffer and Glimsdal, Sondre and Jiao, Lei and Goodwin, Morten and Omlin, Christian W. and Berge, Geir Thore},
  journal={arXiv preprint arXiv:1905.09688},
  year={2019}
}

@article{abeyrathna2021adaptive,
  title={Adaptive Sparse Representation of Continuous Input for Tsetlin Machines Based on Stochastic Searching on the Line},
  author={Abeyrathna, K. Darshana and Granmo, Ole-Christoffer and Goodwin, Morten},
  journal={Electronics},
  volume={10},
  number={17},
  pages={2107},
  year={2021},
  publisher={MDPI},
  doi={10.3390/electronics10172107}
}

@article{jeeru2025mixctme,
  title={MixCTME: A Mixture of Convolutional Tsetlin Machine Experts Using Diverse Spectrogram Visualizations for Jamming Signal Classification},
  author={Jeeru, Sindhusha and Omslandseter, Rebekka Olsson and Andersen, Per-Arne and Morrison, Aiden James and Sokolova, Nadezda and Granmo, Ole-Christoffer and Jiao, Lei},
  journal={IEEE Internet of Things Journal},
  year={2025},
  doi={10.1109/JIOT.2025.3606366}
}

\end{document}